\documentclass[preprint,12pt]{elsarticle}

\usepackage{amssymb}

\usepackage{tabularx}
\usepackage{booktabs}
\usepackage{amsmath}
\usepackage{multirow}
\usepackage{booktabs}
\usepackage{caption}
\usepackage[table,xcdraw]{xcolor}

\usepackage{lineno}
\usepackage{setspace}

\journal{Computers and Electronics in Agriculture}

\begin{document}


\begin{frontmatter}



\title{Evaluating ROCKET and Catch22 features for calf behaviour classification from accelerometer data using Machine Learning models
}


\author[inst1,inst3]{Oshana Dissanayake}
\author[inst3,inst4,inst5]{Sarah E. McPherson}
\author[inst2]{Joseph Allyndrée}
\author[inst3,inst4]{Emer Kennedy}
\author[inst1]{P\'{a}draig Cunningham}
\author[inst1,inst3,inst6]{Lucile Riaboff}

\affiliation[inst1]{organization={School of Computer Science, University College Dublin},
            country={Ireland}}

\affiliation[inst2]{organization={School of Maths and Stats, University College Dublin},
            country={Ireland}}

\affiliation[inst3]{organization={VistaMilk SFI Research Centre},
            country={Ireland}}

\affiliation[inst4]{organization={Teagasc, Animal \& Grassland Research and Innovation Centre},
            addressline={Moorepark, Fermoy}, 
            city={Co. Cork},
            postcode={P61C997}, 
            country={Ireland}}

\affiliation[inst5]{organization={Animal Production Systems Group, Wageningen University \& Research},
            city={Wageningen},
            country={The Netherlands}}

\affiliation[inst6]{organization={GenPhySE, Université de Toulouse},
            city={INRAE, ENVT},
            postcode={31326}, 
            state={Castanet-Tolosan},
            country={France}}

\begin{abstract}
Monitoring calf behaviour continuously would be beneficial to identify routine practices (e.g., weaning, transport, dehorning, etc.) that impact calf welfare in dairy farms. In that regard, accelerometer data collected from neck collars can be used along with Machine Learning models to classify calf behaviour automatically. However, further development is needed to classify a broad spectrum of behaviours with good genericity from one animal to another. While Hand-Crafted features are typically used in the field as inputs for Machine Learning models, feature sets designed explicitly for time-series classification problems have been developed in related fields, such as ROCKET and Catch22 features. This study aims to compare the performance of ROCKET and Catch22 features to Hand-Crafted features commonly used in the field. 30 Irish Holstein Friesian and Jersey pre-weaned calves were equipped with an accelerometer sensor for several weeks, and their behaviours were annotated, allowing for 27.4 hours of observation aligned with the accelerometer time-series. Additional time-series were computed from the raw X, Y and Z-axis and split into 3-second time windows. ROCKET, Catch22 and Hand-Crafted features were calculated for each time window, and the dataset was then split into the train, validation and test sets. Each set of features was used to train three Machine Learning models (Random Forest, eXtreme Gradient Boosting, and RidgeClassifierCV) to classify six behaviours indicative of pre-weaned calf welfare (drinking milk, grooming, lying, running, walking and \textit{other}). Models were tuned with the validation set, and the performance of each feature-model combination was evaluated with the test set. The best performance across the three models was obtained with ROCKET [average balanced accuracy $\pm$ standard deviation] (0.70 ± 0.07), followed by Catch22 (0.69 ± 0.05), well ahead of Hand-Crafted (0.65 ± 0.034). The best balanced accuracy (0.77) was obtained with ROCKET and RidgeClassifierCV, followed by Catch22 and Random Forest (0.73). Thus, tailoring these approaches for specific behaviours and contexts will be crucial in advancing precision livestock farming and enhancing animal welfare on a larger scale.
\end{abstract}













\end{frontmatter}


\section{Introduction}
\label{sec:introduction}
Enhancing the welfare of young farm animals through the adoption of suitable practices is likely to improve their performance at different scales. Indeed, the prolonged effects of stress can lead to an exhaustion phase, resulting in decreased performance, an increased risk of disease, and slowed growth. In particular, calves are subjected to many stressful events from their first weeks (dehorning, weaning, transport, social isolation, relocation, etc.). Improving calf welfare is thus highly important to prevent physiological changes and death that may happen due to prolonged exposure to stress \citep{koknaroglu2013animal} but also to bridge the gap between farming and society, as consumers place animal welfare as one of their primary expectations \citep{cardoso2016imagining}. Changes in calf behaviour, such as altered feeding patterns, decreased playing behaviour or social isolation, can signal underlying health concerns or environmental stressors \citep{mahendran2023calf, nikkhah2023understanding}. Therefore, monitoring calf behaviour can be a way to identify stress factors \citep{dissanayake2022identification} and to make recommendations on routine practices that are less stressful to promote calf welfare in farming \citep{mcpherson2022effect}. Monitoring the changes in behaviour at an early stage can also help to detect health issues or discomfort as soon as they arise to ensure rapid intervention by the farmer, thus reducing the use of medications and their associated costs. However, automatic and continuous behaviour monitoring is required for the targeted applications, which is not compatible with human observations that are time-consuming and labour-intensive \citep{penning1983technique}. In that regard, sensors that can monitor livestock behaviour automatically have been increasingly used over the last few years \citep{rushen2012automated, jiang2023precision}. In particular, onboard accelerometer sensors offer a continuous data stream in the three-dimensional plane from which a diverse range of movements and activities can be derived. These sensors are also adaptable and versatile, can function for a very long period, and can be integrated into various devices and systems, explaining their rapid adoption in livestock \citep{martiskainen2009cow, moreau2009use, tatler2018high, chakravarty2019novel, yu2023accelerometer}. For example, \citet{gonzalez2015behavioral} have shown that ruminating can be classified with a sensitivity $>85\%$ and specificity $>90\%$ based on a decision tree classifier. \citet{iqbal2021validation} used a decision tree algorithm to classify data segments into foraging, ruminating, travelling, resting, or \textit{other} behaviours, with an accuracy of $>85\%$. Similarly, \citet{benaissa2019use} have shown that the precision and sensitivity for lying behaviour were above 93\% using a leg-mounted accelerometer sensor with Machine Learning models. 

However, \citet{riaboff2022predicting} highlighted in a comprehensive review that the challenge lies in capturing a large spectrum of behaviours, including those only expressed occasionally. For instance, transitional behaviours such as lying-down and standing-up are not frequently observed and are often poorly predicted, with a sensitivity lower than 70\% \citep{martiskainen2009cow, vazquez2015classification}. Some maintenance behaviours (e.g., urinating, drinking), self-grooming behaviours and social interaction also tend to get accuracies lower than 80\% \citep{lush2018classification, rodriguez2020identifying}. Recently, \citet{hosseininoorbin2021deep} stated again that cattle behaviour classification from accelerometer data can provide excellent performance for a small number of classes, but the performance decreases substantially when more than 5 behaviours are considered. However, both core time-budget behaviours and less frequent behaviours are essential for assessing calf welfare. Indeed, the time spent lying-down is usually increased around the time of the diagnosis of respiratory diseases \citep{duthie2021feeding}. \citet{enriquez2010effects} observed that playing stopped abruptly after the physical separation from the dams, while a peak of seeking and walking behaviours was observed the day after the separation. Fewer self-grooming and more scratching events are recorded in the few hours after dehorning, while more lying-down and standing-up transitions are observed \citep{morisse1995effect}. Furthermore, \citet{riaboff2022predicting} also highlighted concerns regarding the generic nature of the models. While the model performance is usually high when the model is tested with data from animals already used for model training, a significant drop in performance is observed when the model is tested with new animals, due to a lack of similarity between the feature spaces of the training and test sets \citep{rahman2018cattle}. In calves, an extensive range of behaviours, including locomotor-play, self-grooming, ruminating, non-nutritive suckling, nutritive suckling, active lying and non-active lying, has been predicted with an accuracy higher than 90\% using the AdaBoost algorithm. However, all the behavioural observations and raw sensor data from the subjects have been merged according to timestamps to create the datasets \citep{carslake2020machine}. Thus, though cross-validation has been performed in model testing, the genericity of the model over different calves has yet to be assessed. This limitation compromises the deployment on experimental and commercial farms, as the models must be robust regardless of the animal. Therefore, improving (i) the range of behaviours correctly predicted and (ii) the genericity of the models are both necessary to evaluate animal welfare and health status, particularly in calves, which have received less attention than cows despite the issues surrounding the welfare of young livestock. 

The prevailing methodology employed in literature for animal behaviour classification entails the segmentation of the accelerometer signal in fixed-size time windows \citep{riaboff2022predicting}. Hand-Crafted (\textbf{HC}) features are then calculated in each time window to capture the dynamic properties of the time-series, predominantly rooted in the time and frequency domains, encompassing mean, median, motion variation, roll, pitch, and spectral entropy \citep{lush2018classification,  chang2022detection, price2022classifying, dissanayake2022identification, balasso2023uncovering}. These features are then used to train a Machine Learning model, usually RandomForest (\textbf{RF}), Linear Discriminant Analysis, Support Vector Machines, Deep Neural Networks, or eXtreme Gradient Boosting. \citep{riaboff2020development, kleanthous2022survey}. While the performance of a set of models has often been compared \citep{dutta2015dynamic, smith2016behavior, hu2020inclusion}, the impact of the feature sets used to classify behaviours in livestock ruminant has yet to be explored. On the contrary, extensive studies in the time-series domain have already highlighted some sets of features that perform very well for classification problems in different applications. In particular, \citet{lubba2019catch22} have identified a set of 22 non-correlated CAnonical Time-series CHaracteristics (called Catch22) initially derived from a large pool of 4791 features based on their exceptional performance across 93 time-series classification datasets. The Catch22 features include the most efficient features, covering auto-correlation, value distributions, outliers, and fluctuation scaling properties, while minimizing redundancy. Moreover,  \citet{dempster2020rocket, dempster2021minirocket} recently introduced Random Convolutional Kernel Transform (ROCKET) features that surpasses conventional approaches in time-series classification problems. ROCKET uses Random Convolutional Kernels, with kernels of random length, weights, bias, dilation and padding to transform raw time-series into high-dimensional feature representations. In that way, ROCKET captures diverse temporal patterns without domain-specific knowledge to distinguish underlying structures and patterns within the data. Although the importance of individual features within different HC feature sets is frequently documented \citep{lush2018classification, dissanayake2022identification}, the performance of a set of features developed explicitly for time-series classification problems in related fields, such as Catch22 and ROCKET, has never been investigated. Therefore, this study aims to evaluate the effectiveness of Catch22 and ROCKET for classifying calf behaviour from accelerometer data. Especially, we evaluate Catch22 and ROCKET features performance compared to HC features considering the two main limitations highlighted in the literature, i.e., (i) discriminating a large spectrum of pre-weaned calf behaviours, including drinking milk, running, lying, walking, grooming and \textit{other} and (ii) maintaining good performance when the model is applied to new calves than those used for model training.

\section{Materials and Methods}
\label{sec:mnm}
\begin{figure}[htbp]
\centering
\includegraphics[width=0.95\textwidth]
{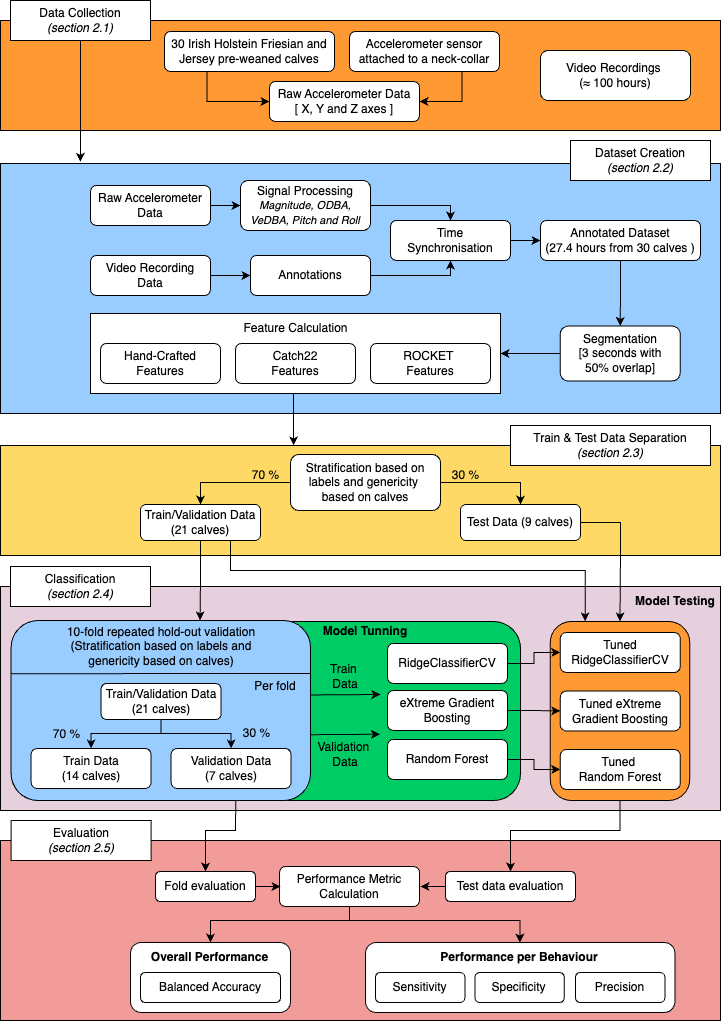}
\captionsetup{labelfont=bf}
\caption{Overview of the methodology applied to compare classification performance depending on the subset of features (HC, Catch22 and ROCKET) and models (RF, XGB, RidgeClassifierCV) and evaluate model genericity across calves.}
\label{fig:pipeline}
\end{figure}

An overview of the methodology applied is described in Figure \ref{fig:pipeline}. Each step is then detailed in the following sections.

\subsection{Data Collection}
\label{sec:data_collection}
\begin{figure}[hbtp]
\centering
\includegraphics[width=0.95\textwidth]{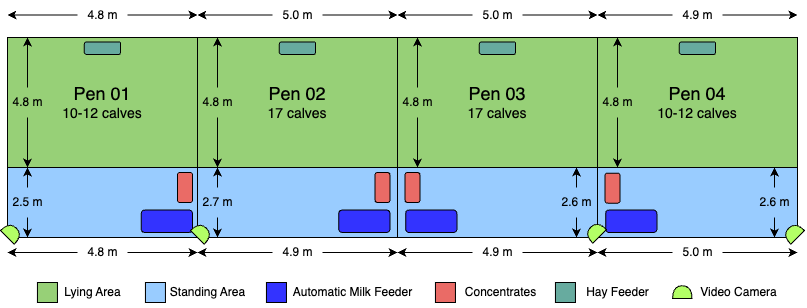}
\captionsetup{labelfont=bf}
\caption{Design of the group pens used after 3-7 days old of age with cameras placement to film the calves in each pen.}
\label{fig:pen-layout}
\end{figure}

The experiment was carried out at Teagasc Moorepark Research Farm (Fermoy, Co. Cork, Ireland; $50^{\circ}07'N$; $8^{\circ}16'W$) from January 21 to April 5, 2022. Ethical approval was obtained from the Teagasc Animal Ethics Committee (TAEC; TAEC2021–319). The trial was carried out in accordance with the European Union (Protection of Animals Used for Scientific Purpose) Regulations 2012 (S.I. No. 543 of 2012). 
47 Irish Holstein Friesian and Jersey pre-weaned calves were utilized for the experiment. The calves were managed according to conventional rearing and management practices \citep{conneely2014effects} at Teagasc Moorepark Research Farm. After calving, the calves and dams were separated within one hour. The calves were moved to a straw-bedded individual pen and were artificially fed with their mother’s colostrum $<2h$ post-birth at a rate of 8.5\% of their birth weight. After receiving colostrum, calves were fed their own dam’s transition milk at a rate of 10\% of their birth weight twice a day for their subsequent five feedings. After transition milk, calves were fed a 2.5L milk replacer (26\% crude protein; Volac Heiferlac Instant, Volac, Hertfordshire, UK) twice daily. Between 3-7 days old, the calves were transferred to a group pen (see Figure \ref{fig:pen-layout}), where they were fed using an automatic milk feeder at a rate of 6 L/calf/day with \textit{ad libitum} access to hay, concentrates, and water. Calves were gradually weaned at 56 days using the automatic feeder.  

\begin{figure}[hbtp]
\centering
\includegraphics[width=0.5\textwidth]{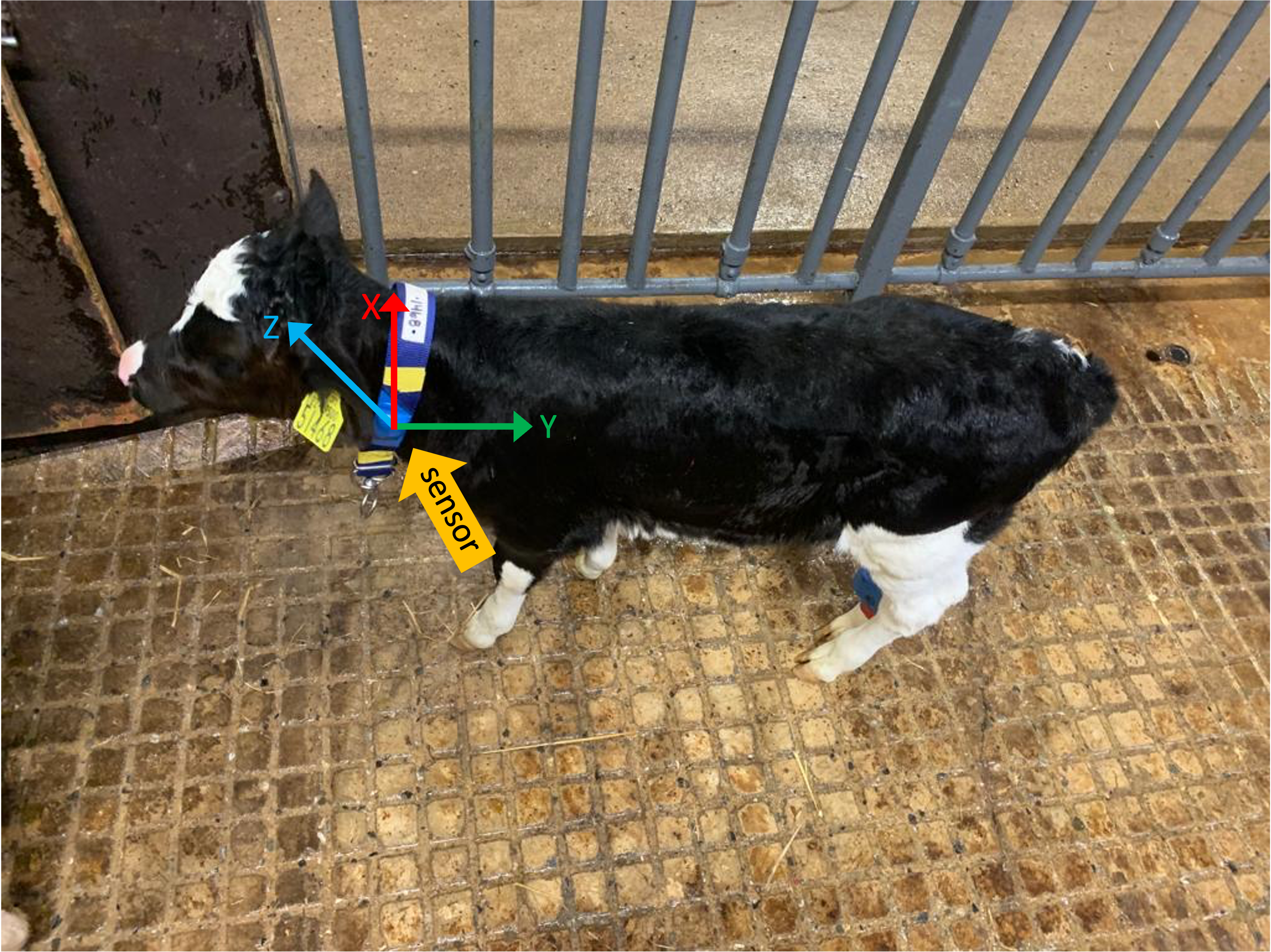}
\captionsetup{labelfont=bf}
\caption{The collar used to attach the accelerometer sensor and the orientation of the three axes. The X-axis detected the top-bottom direction, the Y-axis detected the backwards-forward direction, and the Z-axis detected the left–right direction.}
\label{fig:sensor-orientation}
\end{figure}

Each calf was equipped with a tri-axis accelerometer data logger (Axivity LTD\footnote{Axivity Ltd; https://axivity.com/product/ax3}) fastened to a neck collar starting from one week after birth until 2 months of age (see Figure \ref{fig:sensor-orientation}). The accelerometers were configured with a sensitivity of $\pm 8$g and a sampling rate of 25 Hz (battery life: $30$ days). A NAND flash memory was used to store the data (memory: $512$Mb). The accelerometers were measuring 23×32.5×7.6 mm and weighing $11$g. Each sensor was wrapped in cling film and cotton wool and attached to the collar with vet wrap and insulating tape. The sensors were placed on the left side of the neck in the same orientation for all calves (see Figure \ref{fig:sensor-orientation}).The X-axis detected the top-bottom direction, the Y-axis detected the backwards-forward direction, and the Z-axis detected the left–right direction. The collars were tightly adjusted, and a $13$g metal ring was added to prevent them from moving from the designated side (Figure \ref{fig:sensor-orientation}). For the next ten weeks, collars were taken off every two weeks to retrieve data and replenish the battery. A set of DVRs (4-channel and 8-channel Hikvision\footnote{Equicom Ltd; https://www.equicom.ie/8-channel-dvr-recorder-for-cctv-cameras.html} 1080p) and four 8Mp Dome\footnote{Equicom Ltd; https://www.equicom.ie/1080p-mini-dome-cctv-camera.html} CCTV cameras mounted in each pen were employed in addition to the accelerometer data collection to record videos of calves (see Figure \ref{fig:pen-layout}). 2092 hours of videos were extracted at the end of the trial. 

A subset of 30 calves out of the 47 who demonstrated a wide range of behaviours within the selected videos were selected to create the dataset for this study. Roughly 3 hours of video time were inspected per calf out of multiple videos. This was done to ensure that the selected videos included a wide range of behaviours displayed by each calf across different situations. Labelling was done with the Behavioural Observation Research Interactive Software (BORIS) \citep{friard2016boris} using an exhaustive ethogram with 24 behaviours based on \citet{barry2019development} after adaptation to this experiment. Annotations were carried out by  three observers, ensuring that all 30 animals were observed for at least 15 minutes by at least one observer. The three observers annotated a one-hour video for one calf to measure the concordance between annotations every second based on the 24 behaviours (Cohen’s Kappa averaged over the 1-hour video: $0.72 +/- 0.01$). A total of 27.4 hours of observation has been performed over the 30 calves (age: $23.7 +/- 10.7$ days). Finally, five behaviours indicative of calf welfare each representing more that 1.7\% of the whole dataset were kept in the study, i.e., lying (42.1\%), drinking milk (9.23\%), grooming (4.29\%), running (2.24\%) and walking (1.71\%). All the other behaviours were merged into a class \textit{other} (40.42\%) and retained in the rest of the process. The definitions of each behaviour are listed in Table \ref{tab:behaviour-definitions} and, the amount of data collected for each behaviour is detailed in Table \ref{tab:behaviour-summary}.

The accelerometer time-series were aligned with the observations after synchronizing the accelerometer timestamps with the video timestamps. This was achieved using an external reference clock used during the data collection to make the correspondence between the accelerometer and camera timestamps.

\begin{table}[hbtp]
\captionsetup{labelfont=bf}
\caption{Definition of the different behaviours \citep{barry2019development} in the ethogram used for that study after merging the least observed behaviours ($<$1.7 \%) in the \textit{other} class.}
\label{tab:behaviour-definitions}
\resizebox{\textwidth}{!}{
\begin{tabular}{@{}ll@{}}
\toprule
\multicolumn{1}{c}{Behaviour} & \multicolumn{1}{c}{Definition}                                         \\ \midrule
Drinking milk              & Calf is drinking milk from the milk feeder.                          \\ \midrule
Grooming                     & Calf uses tongue to repeatedly lick own back, side, leg, tail areas. \\ \midrule
Lying & \begin{tabular}[c]{@{}l@{}}Calf is resting either sternally or laterally with all four legs \\ hunched close to body either awake or asleep.\end{tabular} \\ \midrule
Running                      & Calf is running (play / not-play).                                                     \\ \midrule
Walking                      & Calf is walking or shuffling about.                                  \\ \midrule
\textit{Other} & \begin{tabular}[c]{@{}l@{}}A collection of 19 other behaviours including rising, lying-down, \\ social interaction, play etc.\end{tabular}                   \\ \bottomrule
\end{tabular}
}
\end{table}

\begin{table}[hbtp]
\captionsetup{labelfont=bf}
\caption{Summary of data collected: Duration of observation, number of sequences associated with each behaviour, and number of calves on which each behaviour was observed.}
\label{tab:behaviour-summary}
\resizebox{\textwidth}{!}{
\begin{tabular}{@{}rrrr@{}}
\toprule
Behaviour       & Total Duration (minutes) & Number of sequences & Number of calves \\ \midrule
Drinking\ milk & 273.6                  & 169                 & 27               \\
Grooming        & 127.35                   & 334                 & 29               \\
Lying           & 1248.35                  & 120                 & 27               \\
Running         & 66.45                   & 608                 & 24               \\
Walking         & 50.75                   & 561                 & 30               \\
\textit{Other}           & 1198.6                  & 2636                & 30               \\ \bottomrule
\end{tabular}}
\end{table}

\subsection{Creation of the 3 features datasets}
\label{sec:feature_dataset_creation}
The computational workflow was executed on an Intel Xeon E-2378G CPU (2.80GHz and based on x86\_64 architecture) with 16 CPUs, eight cores per socket, and two threads per core with a Matrox G200eW3 GPU.

\subsubsection{Accelerometer time-series calculation}
\label{sec:acc_ts_calc}
A set of time-series typically calculated in the literature was derived from the X, Y, and Z axes, as recommended in \citet{riaboff2022predicting}. The magnitude of the acceleration was first computed from the raw X, Y and Z axes readings of the accelerometer, removing the gravity component of the acceleration \citep{vanHees2013, da2014physical} (equation \ref{eq:magnitude}). A $6^{th}$-order Butterworth high-pass filter with a cut-off frequency of 0.3 Hz \citep{smith2016behavior} was applied to isolate the acceleration component linked to the movement of the animal's body (dynamic acceleration). Overall Dynamic Body Acceleration (ODBA) \citep{wilson2008prying, versluijs2023classification} and Vectorial Dynamic Body Acceleration (VeDBA) \citep{walker2015prying, versluijs2023classification} were computed from the dynamic acceleration according to the equations \ref{eq:ODBA} and \ref{eq:VeDBA}, respectively. A $6^{th}$-order Butterworth low-pass filter with a cut-off frequency of 0.3 Hz was applied to isolate the acceleration component linked to gravity (static acceleration). Pitch \citep{walker2015prying, versluijs2023classification} and roll \citep{walker2015prying, versluijs2023classification} were computed from the static component according to equations \ref{eq:pitch} and \ref{eq:roll}, respectively. 

\begin{equation} \label{eq:magnitude}
        magnitude = \sqrt{(A_x^2 + A_y^2 + A_z^2)} - 1g
\end{equation}

\begin{equation} \label{eq:ODBA}
        ODBA = |D_x| + |D_y| + |D_z|
\end{equation}

\begin{equation} \label{eq:VeDBA}
        VeDBA = \sqrt{(D_x^2 + D_y^2 + D_z^2)}
\end{equation}

\begin{equation} \label{eq:pitch}
        pitch = \arctan(S_z/(S_y^2 + S_x^2))
\end{equation}

\begin{equation} \label{eq:roll}
        roll = \arctan(S_y/(S_z^2 + S_x^2))
\end{equation}

where $A_x, A_y$ and $A_z$ represents the X, Y and Z axes reading of the accelerometer, $D_x, D_y$, and $D_z$ represent the dynamic component of the acceleration and $S_x, S_y$ and $S_z$ the static component of the acceleration.

\subsubsection{Segmentation into fixed time-windows}
\label{sec:segmentation}
Each time-series was split into a 3 seconds-window to (i) meet the [3 - 30 seconds] range as recommended in the literature to get enough signal information within a window \citep{riaboff2022predicting} and (ii) get a time-window short enough to adapt to the brief and changing behaviours of pre-weaned calves. As the literature recommends, a 50\% overlap has been applied between each window \citep{riaboff2022predicting}. It should be noted that the best window size and overlap combination was not investigated in this study as our main objective was to compare the performance of different subsets of features along with Machine Learning models rather than optimising a methodological framework.

\subsubsection{Feature Calculation}
\label{sec:feature_calc}
ROCKET, Catch22, and Hand-Crafted (HC) features were calculated from each 3-seconds window across the 8 time-series: X, Y, and Z axes, magnitude, OBDA, VeDBA, pitch and roll.

\begin{enumerate}
    \itemsep1em
    \item \textbf{Hand-Crafted features} \\
    HC features calculated in livestock ruminant classification behaviour from accelerometer data are usually extracted  in the time domain and frequency domain using the Fourier Transform. Those features provide information on the motion intensity (e.g., median, quartiles, maximum, motion variation from ODBA and VeDBA), the orientation of the animal’s body (e.g., pitch and roll), the shape of the signal distribution (e.g., skewness and kurtosis) and the physical properties of the movement (periodic, stochastic, etc.) (e.g., spectral entropy, fundamental frequency, etc.) \citep{riaboff2022predicting}. In this study, we calculated mean, median, minimum, maximum, standard deviation, first quartile, third quartile, skewness, kurtosis, entropy and motion variation. Across the 8 time-series, this led to a set of 88 HC features typically calculated in the field, providing information on the four categories listed above. This feature set was used as the reference to evaluate the potential additional value of Catch22 and ROCKET. Features, extraction domain, equations and references are displayed in Table \ref{tab:hc-feature-definitions}. 

    \begin{table}[hbtp]
    \captionsetup{labelfont=bf}
    \caption{Hand-Craft features calculated as the reference: Equations, domain and references.}
    \label{tab:hc-feature-definitions}
    \resizebox{\textwidth}{!}{
    \begin{tabular}{@{}lcll@{}}
    \toprule
    \multicolumn{1}{c}{\textbf{Feature}} &
      \textbf{Domain} &
      \multicolumn{1}{c}{\textbf{Equation}} &
      \multicolumn{1}{c}{\textbf{Reference}} \\ \midrule
    Mean &
      Time &
      \begin{tabular}[c]{@{}l@{}}$\text{Mean} = \frac{\sum_{1}^{N} X_i}{N}$\\ N = Total number of observations.\end{tabular} &
      \begin{tabular}[c]{@{}l@{}}\citep{preece2008comparison}\\ \citep{drover2017faller}\\ \citep{tatler2018high}\end{tabular} \\
      \midrule
    Median &
      Time &
      \begin{tabular}[c]{@{}l@{}}$M = \begin{cases} \\ X_{\frac{n+1}{2}}, & \text{if } n \text{ is odd} \\\\ \frac{X_{\frac{n}{2}} + X_{\frac{n}{2} + 1}}{2}, & \text{if } n \text{ is even}\\ \end{cases}$\\ X = Data points in the sorted list.\end{tabular} &
      \begin{tabular}[c]{@{}l@{}}\citep{preece2008comparison}\\ \citep{figo2010preprocessing} \\ \citep{fida2015pre}\end{tabular} \\
      \midrule
    Minimum &
      Time &
      \begin{tabular}[c]{@{}l@{}}$\text{Min}(X) = \min(X_1, X_2, \ldots, X_n)$\\ X = Considered set of data points.\end{tabular} &
      \begin{tabular}[c]{@{}l@{}}\citep{figo2010preprocessing}\\ \citep{barwick2018categorising}\end{tabular} \\
      \midrule
    Maximum &
      Time &
      \begin{tabular}[c]{@{}l@{}}$\text{Min}(X) = \max(X_1, X_2, \ldots, X_n)$\\ X = Considered set of data points.\end{tabular} &
      \begin{tabular}[c]{@{}l@{}}\citep{figo2010preprocessing}\\ \citep{barwick2018categorising}\end{tabular} \\
      \midrule
    \begin{tabular}[c]{@{}l@{}}Standard\\ Deviation\end{tabular} &
      Time &
      \begin{tabular}[c]{@{}l@{}}$\sigma = \sqrt{\frac{\sum (x_i-\mu )^2}{N}}$\\ N = size of the data segment. \\ $x_i$ = Individual observation in the data set.\\ $\mu$ = mean of the data segment.\end{tabular} &
      \begin{tabular}[c]{@{}l@{}}\citep{preece2008comparison}\\ \citep{figo2010preprocessing}\\ \citep{bersch2014sensor}\\ \citep{drover2017faller}\end{tabular} \\
      \midrule
    \begin{tabular}[c]{@{}l@{}}First\\ Quartile\end{tabular} &
      Time &
      \begin{tabular}[c]{@{}l@{}}$Q1 = \frac{N+1}{4}$\\ N = Total number of observations in the data set.\end{tabular} &
      \begin{tabular}[c]{@{}l@{}}\citep{preece2008comparison}\\ \citep{zdravevski2017automatic}\\ \citep{fan2019feature}\end{tabular} \\
      \midrule
    \begin{tabular}[c]{@{}l@{}}Third\\ Quartile\end{tabular} &
      Time &
      \begin{tabular}[c]{@{}l@{}}$Q3 = \frac{3}{4}(N+1)$\\ N = Total number of observations in the data set.\end{tabular} &
      \begin{tabular}[c]{@{}l@{}}\citep{preece2008comparison}\\ \citep{zdravevski2017automatic}\\ \citep{fan2019feature}\end{tabular} \\
      \midrule
    \begin{tabular}[c]{@{}l@{}}Spectral\\ Entropy\end{tabular} &
      Frequency &
      \begin{tabular}[c]{@{}l@{}}$H(s, sf) = \sum_{f=0}^{f_s/2}P(f)log_2[P(f)]$ \\ Where P is the normalised power spectral density, \\ and $f_s$ is the sampling frequency.\end{tabular} &
      \begin{tabular}[c]{@{}l@{}}\citep{preece2008comparison}\\ \citep{riaboff2020development}\\ \citep{aziz2021machine}\\ \citep{dissanayake2022identification}\end{tabular} \\
      \midrule
    \begin{tabular}[c]{@{}l@{}}Motion \\ Variation\end{tabular} &
      Time &
      \begin{tabular}[c]{@{}l@{}}$MV = \frac{1}{M}(\sum_{i=1}^{M-1}|a_{x,i+1} - a_{x,i}| +$ \\ \qquad \qquad \: $\sum_{i=1}^{M-1}|a_{y,i+1} - a_{y,i}| +$ \\ \qquad \qquad \: $\sum_{i=1}^{M-1}|a_{z,i+1} - a_{z,i}|)$\end{tabular} &
      \begin{tabular}[c]{@{}l@{}}\citep{riaboff2020development}\\ \citep{fogarty2020behaviour}\\ \citep{dissanayake2022identification}\end{tabular} \\ 
      \midrule
  Skewness &
      Time &
      \begin{tabular}[c]{@{}l@{}}$\gamma = \frac{1}{N} \sum_{i=1}^N \left( \frac{Y_i - \mu}{\sigma} \right)^3$ \\ \\
      $\gamma$ = skewness\\
      N = number of variables in the distribution \\
      $\mu$ = mean of the distribution\\
      $\sigma$ = standard deviation\end{tabular} &
      \begin{tabular}[c]{@{}l@{}}\citep{alzubi2014human}\\ \citep{hounslow2019assessing} \\ \citep{cabezas2022analysis}\end{tabular} \\
      \midrule 
  Kurtosis &
      Time &
      \begin{tabular}[c]{@{}l@{}}$\beta = \frac{\frac{1}{N} \sum_{i=1}^N (Y_i - \mu)^4}{V^2}$\\ \\
      $\beta$ = kurtosis\\
      N = number of variables in the distribution \\
      $\mu$ = mean of the distribution\\
      V = variance of the dataset\end{tabular} &
      \begin{tabular}[c]{@{}l@{}}\citep{alzubi2014human}\\ \citep{hounslow2019assessing} \\ \citep{cabezas2022analysis}\end{tabular} \\
      \bottomrule
    \end{tabular}}
    \end{table}

    \item \textbf{Catch22 features} \\
    Catch22 are 22 CAnonical Time-series CHaracteristics derived from a large pool of 4791 features that showed exceptional performance across 93 time-series classification datasets. This subset was also selected to minimize redundancy between features, thus providing complementary information on the time-series while reducing computational expenses \citep{lubba2019catch22}. The Catch22 features cover auto-correlation, value distributions, outliers, and fluctuation scaling properties. This set can be increased to 24 features by including mean and standard deviation to include the location and spread of the raw time-series distribution in the classification process. As that information may be necessary for the time-windows classification into calf behaviour, the set of 24 features was calculated, leading to a subset of 192 features across the 8 time-series. For the remainder of the paper, we are keeping the name \textit{Catch22} as soon as we refer to that features set. Feature names and their description are displayed in Table \ref{tab:catch22-feature-definitions}. 

    \begin{table}[hbtp]
    \captionsetup{labelfont=bf}
    \caption{Catch22 features for time-series classification problem \citep{lubba2019catch22}.}
    \label{tab:catch22-feature-definitions}
    \resizebox{\textwidth}{!}{
    \begin{tabular}{@{}ll@{}}
    \toprule
    Feature name                                & Description                                                               \\ \midrule
    DN\_HistogramMode\_5                        & Mode of z-scored distribution (5-bin histogram)                           \\
    DN\_HistogramMode\_10                       & Mode of z-scored distribution (10-bin histogram)                          \\
    SB\_BinaryStats\_mean\_longstretch1         & Longest period of consecutive values above the mean                       \\
    DN\_OutlierInclude\_p\_001\_mdrmd           & Time intervals between successive extreme events above the mean           \\
    DN\_OutlierInclude\_n\_001\_mdrmd           & Time intervals between successive extreme events below the mean           \\
    CO\_f1ecac                                  & First 1/e crossing of autocorrelation function                            \\
    CO\_FirstMin\_ac                            & First minimum of autocorrelation function                                 \\
    SP\_Summaries\_welch\_rect\_area\_5\_1              & Total power in lowest fifth of frequencies in the Fourier power spectrum                 \\
    SP\_Summaries\_welch\_rect\_centroid        & Centroid of the Fourier power spectrum                                    \\
    FC\_LocalSimple\_mean3\_stderr              & Mean error from a rolling 3-sample mean forecasting                       \\
    CO\_trev\_1\_num                            & Time-reversibility statistic, $(xt+1 - xt)3t$                               \\
    CO\_HistogramAMI\_even\_2\_5                & Automutual information, m = 2, $\tau$ = 5                                      \\
    IN\_AutoMutualInfoStats\_40\_gaussian\_fmmi & First minimum of the automutual information function                      \\
    MD\_hrv\_classic\_pnn40                     & Proportion of successive differences exceeding $0.04\sigma$ \citep{mietus2002pnnx} \\
    SB\_BinaryStats\_diff\_longstretch0         & Longest period of successive incremental decreases                        \\
    SB\_MotifThree\_quantile\_hh                        & Shannon entropy of two successive letters in equiprobable 3-letter symbolization         \\
    FC\_LocalSimple\_mean1\_tauresrat           & Change in correlation length after iterative differencing                 \\
    CO\_Embed2\_Dist\_tau\_d\_expfit\_meandiff          & Exponential fit to successive distances in 2-d embedding space                           \\
    SC\_FluctAnal\_2\_dfa\_50\_1\_2\_logi\_prop\_r1     & Proportion of slower timescale fluctuations that scale with DFA (50\% sampling)          \\
    SC\_FluctAnal\_2\_rsrangefit\_50\_1\_logi\_prop\_r1 & Proportion of slower timescale fluctuations that scale with linearly rescaled range fits \\
    SB\_TransitionMatrix\_3ac\_sumdiagcov               & Trace of covariance of transition matrix between symbols in 3-letter alphabet            \\
    PD\_PeriodicityWang\_th0\_01                & Periodicity measure of \text{\citep{wang2007structure}}          \\ \bottomrule
    \end{tabular}
    }
    \end{table}

    \item \textbf{ROCKET features} \\
    ROCKET (RandOm Convolutional KErnel Transform) is a pioneering technique in time-series classification, known for its efficiency and efficacy in features extraction. ROCKET uses Random Convolutional Kernels, where kernels have random lengths, weights, biases, dilations, and paddings, to transform raw time-series into high-dimensional feature representations. ROCKET features are specifically the maximum and proportion of positive values resulting from each convolution, meaning that for k kernels, 2k features per time-series are produced. Using random kernels allows to capture diverse temporal patterns without domain-specific knowledge. For this study, we have utilized \textit{miniROCKET} which leverages a more streamlined set of convolutional kernels, focusing on a predefined subset of kernel parameters rather than the exhaustive randomness of ROCKET. This refinement allows miniROCKET to achieve similar or even superior classification performance with a significantly reduced computational footprint \citep{dempster2021minirocket}. The default number of kernels (10000) was used during the feature generation in that study. ROCKET features were calculated from each of the 8 time-series, leading to a set of 9996 features. 
\end{enumerate}

\subsection{Partitioning the dataset into training and testing calf-independent sets}
\label{sec:train_test_split}
The dataset was split into a 70:30 calf ratio (Figure \ref{fig:pipeline}): Out of 30 calves, 21 were chosen for the training set, and 9 were chosen for the test set. This split ensures that the calves used for testing the model have not been used for model training to evaluate the model genericity. Furthermore, stratification has been applied to the annotated behaviours to maintain a consistent proportional distribution between the train and test set. The optimal split used in the rest of the study is the one that minimises the mean of the differences in the proportions of each of the annotated behaviours between the training and testing sets. 

\subsection{Modelling with Machine Learning models}
\label{sec:ML_models}
Three Machine Learning models were used along with each feature set to (i) conclude on the average performance of the feature sets across several models and (ii) investigate whether some feature set-model combinations perform better than others. eXtreme Gradient Boosting (\textbf{XGB}), RandomForest \textbf{(RF)} and RidgeClassifierCV \textbf{(RCV)} algorithms were used, considering their high performance in classification problems in various domains.

\begin{enumerate}
    \item \textbf{eXtreme Gradient Boosting} \\
    The eXtreme Gradient Boosting (XGB) algorithm is an ensemble Machine Learning technique that uses gradient boosting techniques to improve model accuracy \citep{friedman2001greedy}. It constructs decision trees sequentially, with each tree correcting previous errors to enhance model accuracy. The algorithm uses gradient boosting, where each tree is trained using the gradient of the loss function, minimizing the difference between predicted and actual values. XGB is highly versatile, capable of handling sparse data and working with various data formats. It also incorporates techniques to prevent overfitting, such as regularization terms in the objective function. XGB offers efficient scalability and parallel processing capabilities, significantly speeding up computation time, especially with large datasets. Below are a few most prominent hyper-parameters \citep{chen2016xgboost}. The values tested for each hyperparameter in this study are shown in Table \ref{tab:hyper-params}.

    \begin{itemize}
        \item \textbf{eta (learning rate): }Determines the step size at each iteration while moving toward a minimum of the loss function. A lower value makes the model more robust at the cost of slower computation.

        \item \textbf{max depth:} Sets the maximum depth of a tree. Increasing this value will make the model more complex and likely to overfit.

        \item \textbf{n\_estimators: }Number of trees to be built. More trees can increase accuracy but also computation time.

        \item \textbf{gamma: }Minimum loss reduction required to make a further partition on a leaf node of the tree. The larger the value, the more conservative the algorithm will be.

        \item \textbf{lambda (reg lambda) and alpha (reg alpha):} These are L2 and L1 regularization terms on weights, respectively, and can be used to handle overfitting.

        \item \textbf{scale pos weight: }Controls the balance of positive and negative weights, useful for unbalanced classes.
    \end{itemize}

    \item \textbf{Random Forest} \\
    The Random Forest algorithm is a versatile ensemble learning method for classification tasks. It constructs multiple decision trees during training and outputs the mode of classes or mean prediction of the individual trees. Randomness is introduced through bootstrap sampling and a random subset of features at each node, making the model more robust and preventing overfitting. The algorithm can handle large datasets effectively with higher dimensionality, performs well with categorical and continuous variables, and can handle missing values \citep{breiman2001random}. Below are a few most prominent hyper-parameters based on \citep{scikit-learn} and the tested values are shown in Table \ref{tab:hyper-params}.

    \begin{itemize}
        \item \textbf{n\_estimators: }The number of trees in the forest.
        \item \textbf{criterion: }The function to measure the quality of a split.
        \item \textbf{max depth: }The maximum depth of the tree.
        \item \textbf{min samples split: }The minimum number of samples required to split an internal node. Higher values prevent creating trees that are too complex and overfitting.
        \item \textbf{max features: }The number of features to consider when looking for the best split. It helps in adding randomness to the model.
        \item \textbf{class weight: }Weights associated with classes. It is particularly useful in scenarios where classes are imbalanced, as it can influence the decision trees in the forest towards the minority class.
    \end{itemize}

    \item \textbf{RidgeClassifierCV} \\
    RidgeClassifierCV is useful in large-variable scenarios and aims to prevent overfitting while maintaining a balance between bias and variance. It is efficient for high-dimensional datasets and is commonly used in problems requiring interpretability and prediction accuracy. RidgeClassifierCV \citep{scikit-learn} is an extension of the RidgeClassifier based on the theory of Ridge Regression \citep{hoerl1970ridge}, which is an extension of linear regression that uses cross-validation to determine the optimal regularization parameter. This method splits the dataset into subsets and evaluates the model’s performance for different regularization parameter (alpha) values. The \textbf{CV} stands for cross-validation, which helps to find the alpha that yields the best generalization performance. Following are a few important hyperparameters of RidgeClassifierCV \citep{scikit-learn} and the tested values are shown in Table \ref{tab:hyper-params}.

    \begin{itemize}
        \item \textbf{alphas: }Determines the regularization strength. Regularization improves the conditioning of the problem and reduces the variance of the estimates. Larger values promotes stronger regularization.
        \item \textbf{fit\_intercept: }Determines whether a bias (intercept) term is added to the decision function, allowing the model to fit data that is not centered around zero.
        \item \textbf{cv: }Determines the cross-validation splitting strategy.
        \item \textbf{class\_weight: }Weights associated with classes. It is particularly useful in scenarios where classes are imbalanced.
    \end{itemize}
\end{enumerate}

\begin{table}[hbtp]
\captionsetup{labelfont=bf}
\caption{Tested Classifiers and hyper-parameters.}
\label{tab:hyper-params}
\resizebox{\textwidth}{!}{
\begin{tabular}{lll}
\hline
\textbf{Classifier} &
  \textbf{Hyperparameters} &
  \textbf{Values Tested} \\ \hline
XGBoost &
  \begin{tabular}[c]{@{}l@{}}n\_estimators\\ eta\\ gamma\\ max\_depth\\ class\_weight\end{tabular} &
  \begin{tabular}[c]{@{}l@{}}100, 200\\ 0, 0.5, 1\\ 0, 5, 10\\ None, 0, 10\\ None, balanced\end{tabular} \\ \hline
Random Forest &
  \begin{tabular}[c]{@{}l@{}}n\_estimators\\ max\_depth\\ min\_samples\_split\\ max\_features\\ criterion\\ class\_weight\end{tabular} &
  \begin{tabular}[c]{@{}l@{}}100, 200\\ None, 10\\ 2, 5\\ None, log2, sqrt\\ gini, entropy\\ None, balanced\end{tabular} \\ \hline
RidgeClassifierCV &
  \begin{tabular}[c]{@{}l@{}}fit\_intercept\\ class\_weight\\ alphas\end{tabular} &
  \begin{tabular}[c]{@{}l@{}}True, False\\ None, balanced\\ np.logspace(-3,3,100), np.logspace(-1,10,100)\end{tabular} \\ \hline
\end{tabular}}
\end{table}

\subsection{Model tuning and evaluation for each feature set-model combination}
\label{sec:model_tune_evaluation}
The Machine Learning models were developed using scikit-learn ($v.1.2.2$) \citep{scikit-learn}, XGBoost ($v.1.6.1$) \citep{chen2016xgboost}, and skTime ($v.0.24.0$) \citep{loning2019sktime} available under Python $v.3.9.7.$ For each of the 3 sets of features, the 3 Machine Learning models were tuned using the training data. A grid search was used to identify the best hyper-parameters for each model along with each set of features (see Table \ref{tab:best-hyper-params}). For that purpose, each model was trained with one of the combinations of hyperparameters in the grid search using 14 calves from the 21 available in the training dataset. The model was then evaluated using a validation dataset made up of the 7 remaining calves in the training set. The process has been iterated 10 times (see Figure \ref{fig:pipeline}). For each feature set and algorithm, the model that achieved the best balanced accuracy (BA; see equation \ref{eq:BA}) on average over the 10 iterations on the validation dataset was selected for the rest of the study (tuned model). For each set of features, the performances of each of the 3 tuned Machine Learning models were calculated using the BA as a global metric of performance and using the sensitivity, specificity and precision as a performance metric per behaviour (see equation \ref{eq:sensitivity}, \ref{eq:specificity} and \ref{eq:precision}, respectively).

\begin{table}[hbtp]
\captionsetup{labelfont=bf}
\caption{Best performing hyper-parameters identified through the grid search for each feature set-model combination.}
\vspace{3pt}
\label{tab:best-hyper-params}
\resizebox{\textwidth}{!}{
\begin{tabular}{@{}lllll@{}}
\toprule
\textbf{Model} & \textbf{Hyper-parameters} & \textbf{Hand-Crafted} & \textbf{Catch22} & \textbf{ROCKET} \\ \midrule
\multirow{5}{*}{XGB}               & class weight        & None                   & None                   & None                   \\
                                   & eta                 & 0.5                    & 1                      & 1                      \\
                                   & gamma               & 0                      & 0                      & 0                      \\
                                   & max\_depth          & 10                     & 6                      & 6                      \\
                                   & n\_estimators       & 200                    & 200                    & 200                    \\\midrule
\multirow{6}{*}{RF}                & class weight        & balanced               & balanced               & balanced               \\
                                   & criterion           & gini                   & entropy                & gini                   \\
                                   & max\_depth          & 10                     & sqrt                   & sqrt                   \\
                                   & max\_features       & log2                   & 10                     & 10                     \\
                                   & min\_samples\_split & 5                      & 5                      & 2                      \\
                                   & n\_estimators       & 200                    & 200                    & 200                    \\\midrule
\multirow{3}{*}{RidgeClassifierCV} & class\_weight       & balanced               & balanced               & balanced               \\
                                   & fit\_intercept      & False                  & True                   & False                  \\
                                   & alphas              & np.logspace(-1,10,100) & np.logspace(-1,10,100) & np.logspace(-1,10,100) \\ \midrule
\end{tabular}}
\end{table}

\begin{equation} \label{eq:BA}
    \text{Balanced Accuracy} = \frac{1}{2}\left(\frac{TP}{TP + FN} + \frac{TN}{TN + FP}\right)
\end{equation}

\begin{equation} \label{eq:sensitivity}
    \text{Sensitivity} = \frac{TP}{TP + FN}
\end{equation}

\begin{equation} \label{eq:specificity}
    \text{Specificity} = \frac{TN}{TN + FP}
\end{equation}

\begin{equation} \label{eq:precision}
    \text{Precision} = \frac{TP}{TP + FP}
\end{equation}

Where \textbf{TP} (True Positive) is the number of time-windows where the behaviour of interest was observed and correctly predicted; \textbf{FN} (False Negative) is the number of time-windows where the behaviour of interest was observed but another behaviour was predicted, \textbf{FP} (False Positive) is the number of time-windows where the behaviour of interest was predicted but another behaviour was observed; \textbf{TN} (True Negative) is the number of time-windows where the behaviour of interest was not observed and not predicted.

The Python code utilized for the sections \ref{sec:feature_dataset_creation}, \ref{sec:train_test_split}, \ref{sec:ML_models} and \ref{sec:model_tune_evaluation} is presented through an \textit{open-source Github project} \footnote{https://github.com/Oshana/comp\_n\_ele\_in\_agriculture.git}.

\section{Results}
\label{sec:results}

\subsection{Model tuning with 10-fold cross-validation}
\label{sec:kfold_validation}
\begin{figure}[hbtp]
\centering
\includegraphics[width=0.95\textwidth]{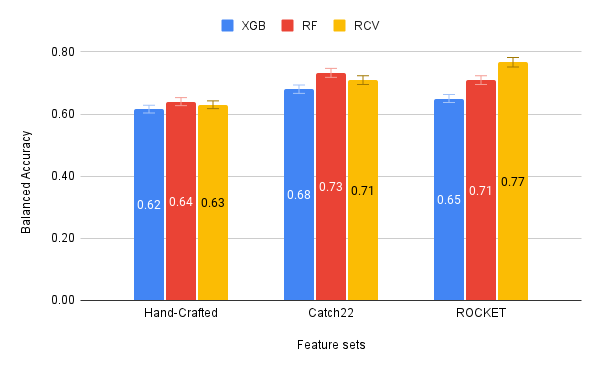}
\captionsetup{labelfont=bf}
\caption{Mean BA with standard-deviation obtained for each 10-folds for every feature set-model combination.}
\label{fig:validation_ba_results}
\end{figure}

The best performing hyperparameters identified through the grid search are detailed in Table \ref{tab:best-hyper-params}, and their respective scores achieved during each 10 fold of the cross-validation for Hand-Crafted, Catch22 and ROCKET feature sets are displayed in Figure \ref{fig:validation_ba_results}.

On average, highest BA across the 10 folds were obtained with ROCKET ($0.70 \pm 0.07$ (\textit{[mean $\pm$ standard-deviation (std)]}) followed by Catch22 ($0.69 \pm 0.05$) and HC features ($0.65 \pm 0.03$) which showed the lowest efficacy.

\subsection{Model performance with the test set}
\label{sec:model_test_performance}

\subsubsection{Overall Performance}
\label{sec:overall_test_performance}

\begin{figure}[hbtp]
\centering
\includegraphics[width=0.95\textwidth]{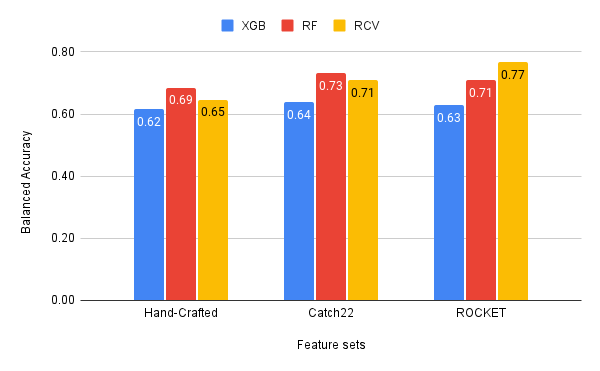}
\captionsetup{labelfont=bf}
\caption{Balanced Accuracy obtained on the calf-independent test-set for every feature set-model combination.}
\label{fig:ba_results}
\end{figure}

The BA scores obtained for each fitted model and set of features on the calf-independent test set are presented in Figure \ref{fig:ba_results}. Overall, the best BA across the 3 Machine Learning models was reached with ROCKET ($0.70 \pm 0.07$ \textit{[$mean \pm std$]}), followed closely by Catch22 ($0.69 \pm 0.05$). BA was substantially lower with HC features ($0.65 \pm 0.03$). 

Variability within the same subset of features has been found depending on the Machine Learning model (Figure \ref{fig:ba_results}). Indeed, ROCKET obtained its highest and lowest BA with RCV and XGB, respectively. Catch22 and HC features reached their best BA with RF and their lowest BA with XGB.

\subsubsection{Performance per behaviour}
\label{sec:perf_per_behaviour}

The best sensitivity, specificity and precision for each behaviour across the 3 Machine Learning models were obtained either with ROCKET or Catch22 features, except for the sensitivity of walking ($0.65 \pm 0.23$ \textit{[$mean \pm std$]}), the specificity of \textit{other} ($0.83 \pm 0.11$) and the precision of running ($0.93 \pm 0.03$) where the best performance were slightly better with HC features (Table \ref{tab:best-per-behav-matrics}; see \ref{sec:appendix} for the complete report of performance). 

\begin{figure}[h]
\centering
\includegraphics[width=1\textwidth]{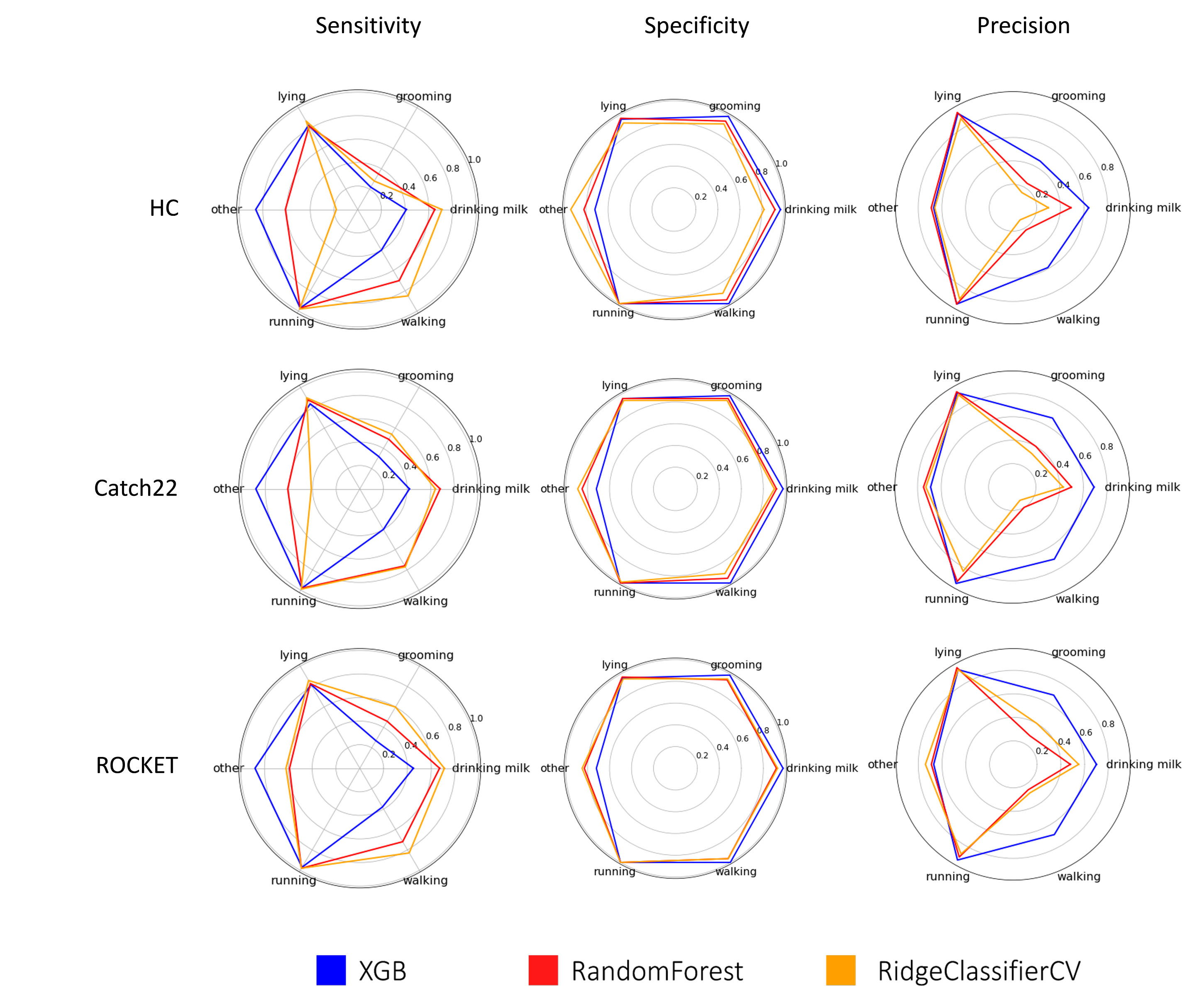}
\captionsetup{labelfont=bf}
\caption{Sensitivity, specificity and precision performance per behaviour for every feature set-model combination.}
\label{fig:perf_per_behav_results}
\end{figure}

Figure \ref{fig:perf_per_behav_results} displays the sensitivity, specificity and precision obtained for each behaviour for each feature set-model combination. Lying and running were correctly predicted regardless of the feature set-model combination, with sensitivity, specificity and precision higher than 0.80. However, some feature set-model combinations are more promising for classifying drinking milk, grooming, walking and \textit{other}. As displayed in Table \ref{tab:best-per-behav-matrics}, drinking milk was better predicted with ROCKET, especially with RCV and XGB. Similarly, grooming achieved its best performance with ROCKET in combination with RCV and XGB and Catch22 in association with XGB. Walking behaviour was best predicted with HC and Catch22, combined with RCV and XGB. The best performance for predicting \textit{other} behaviour class was achieved by combining Catch22 with XGB and RF, ROCKET with XGB, and HC features with RCV.

In the same way as for the BA, the performances per behaviour are highly variable within each set of features, depending on the Machine Learning model (see Table \ref{tab:best-per-behav-matrics} and \ref{sec:appendix} for the full report). The impact of the model on the feature's performance is particularly noticeable for the sensitivity and precision of the behaviours of drinking milk, grooming, walking and \textit{other}. For the HC features, a gap in sensitivity of 0.68, 0.45 and 0.30, respectively, has been found between the highest and the lowest performing model combination for the behaviours \textit{other}, walking and drinking milk, respectively. A difference in precision of 0.34, 0.31 and 0.47 has been obtained between the best and worst model combinations with HC for drinking milk, grooming and walking. For the Catch22 features, substantial differences were also observed depending on the model, with a gap in sensitivity of 0.47, 0.26 and 0.37 for the behaviours \textit{other}, drinking milk and walking, respectively. A high difference in the precision of 0.58, 0.35 and 0.26 for walking, grooming and drinking milk has also been obtained between the best and worst model associated with Catch22. For ROCKET features, a difference in sensitivity of 0.45, 0.33 and 0.26 was found for walking, grooming and drinking milk, respectively, depending on the model. At the same time, a gap in precision of 0.44 and 0.40 was recorded between the best and worst model combinations for the behaviours of walking and grooming. 

\begin{table}[!h]
\captionsetup{labelfont=bf}
\caption{Best sensitivity, specificity and precision achieved per feature set-model combination per behavior.}
\vspace{3pt}
\label{tab:best-per-behav-matrics}
\resizebox{\textwidth}{!}{
\begin{tabular}{@{}|l
>{\columncolor[HTML]{C9DAF8}}l 
>{\columncolor[HTML]{C9DAF8}}l 
>{\columncolor[HTML]{C9DAF8}}l 
>{\columncolor[HTML]{B6D7A8}}l 
>{\columncolor[HTML]{B6D7A8}}l 
>{\columncolor[HTML]{B6D7A8}}l 
>{\columncolor[HTML]{FFE599}}l 
>{\columncolor[HTML]{FFE599}}l 
>{\columncolor[HTML]{FFE599}}l |@{}}
\toprule
\multicolumn{1}{|l|}{} &
  \multicolumn{3}{c|}{\cellcolor[HTML]{C9DAF8}\textbf{HC}} &
  \multicolumn{3}{c|}{\cellcolor[HTML]{B6D7A8}\textbf{Catch22}} &
  \multicolumn{3}{c|}{\cellcolor[HTML]{FFE599}\textbf{ROCKET}} \\ \midrule
\multicolumn{1}{|l|}{Machine learning model} &
  \multicolumn{1}{c|}{\cellcolor[HTML]{C9DAF8}XGB} &
  \multicolumn{1}{c|}{\cellcolor[HTML]{C9DAF8}RF} &
  \multicolumn{1}{c|}{\cellcolor[HTML]{C9DAF8}RCV} &
  \multicolumn{1}{c|}{\cellcolor[HTML]{B6D7A8}XGB} &
  \multicolumn{1}{c|}{\cellcolor[HTML]{B6D7A8}RF} &
  \multicolumn{1}{c|}{\cellcolor[HTML]{B6D7A8}RCV} &
  \multicolumn{1}{c|}{\cellcolor[HTML]{FFE599}XGB} &
  \multicolumn{1}{c|}{\cellcolor[HTML]{FFE599}RF} &
  \multicolumn{1}{c|}{\cellcolor[HTML]{FFE599}RCV} \\ \midrule
\multicolumn{10}{|c|}{\cellcolor[HTML]{7F7F7F}{\color[HTML]{FFFFFF} \textbf{Best sensitivity}}} \\ \midrule
\multicolumn{1}{|l|}{Drinking milk} &
  \multicolumn{1}{l|}{\cellcolor[HTML]{C9DAF8}} &
  \multicolumn{1}{l|}{\cellcolor[HTML]{C9DAF8}} &
  \multicolumn{1}{c|}{\cellcolor[HTML]{34A853}{0.71}} &
  \multicolumn{1}{l|}{\cellcolor[HTML]{B6D7A8}} &
  \multicolumn{1}{l|}{\cellcolor[HTML]{B6D7A8}} &
  \multicolumn{1}{l|}{\cellcolor[HTML]{B6D7A8}} &
  \multicolumn{1}{l|}{\cellcolor[HTML]{FFE599}} &
  \multicolumn{1}{l|}{\cellcolor[HTML]{FFE599}} &
  \multicolumn{1}{c|}{\cellcolor[HTML]{34A853}{ 0.71}} \\ \midrule
\multicolumn{1}{|l|}{Grooming} &
  \multicolumn{1}{l|}{\cellcolor[HTML]{C9DAF8}} &
  \multicolumn{1}{l|}{\cellcolor[HTML]{C9DAF8}} &
  \multicolumn{1}{l|}{\cellcolor[HTML]{C9DAF8}} &
  \multicolumn{1}{l|}{\cellcolor[HTML]{B6D7A8}} &
  \multicolumn{1}{l|}{\cellcolor[HTML]{B6D7A8}} &
  \multicolumn{1}{l|}{\cellcolor[HTML]{B6D7A8}} &
  \multicolumn{1}{l|}{\cellcolor[HTML]{FFE599}} &
  \multicolumn{1}{l|}{\cellcolor[HTML]{FFE599}} &
  \multicolumn{1}{c|}{\cellcolor[HTML]{34A853}{0.60}} \\ \midrule
\multicolumn{1}{|l|}{Lying} &
  \multicolumn{1}{l|}{\cellcolor[HTML]{C9DAF8}} &
  \multicolumn{1}{l|}{\cellcolor[HTML]{C9DAF8}} &
  \multicolumn{1}{l|}{\cellcolor[HTML]{C9DAF8}} &
  \multicolumn{1}{l|}{\cellcolor[HTML]{B6D7A8}} &
  \multicolumn{1}{l|}{\cellcolor[HTML]{B6D7A8}} &
  \multicolumn{1}{c|}{\cellcolor[HTML]{34A853}{0.90}} &
  \multicolumn{1}{l|}{\cellcolor[HTML]{FFE599}} &
  \multicolumn{1}{l|}{\cellcolor[HTML]{FFE599}} &
   \\ \midrule
\multicolumn{1}{|l|}{Running} &
  \multicolumn{1}{l|}{\cellcolor[HTML]{C9DAF8}} &
  \multicolumn{1}{l|}{\cellcolor[HTML]{C9DAF8}} &
  \multicolumn{1}{l|}{\cellcolor[HTML]{C9DAF8}} &
  \multicolumn{1}{l|}{\cellcolor[HTML]{B6D7A8}} &
  \multicolumn{1}{l|}{\cellcolor[HTML]{B6D7A8}} &
  \multicolumn{1}{c|}{\cellcolor[HTML]{34A853}{0.99}} &
  \multicolumn{1}{l|}{\cellcolor[HTML]{FFE599}} &
  \multicolumn{1}{l|}{\cellcolor[HTML]{FFE599}} &
   \\ \midrule
\multicolumn{1}{|l|}{Walking} &
  \multicolumn{1}{l|}{\cellcolor[HTML]{C9DAF8}} &
  \multicolumn{1}{l|}{\cellcolor[HTML]{C9DAF8}} &
  \multicolumn{1}{c|}{\cellcolor[HTML]{34A853}{0.85}} &
  \multicolumn{1}{l|}{\cellcolor[HTML]{B6D7A8}} &
  \multicolumn{1}{l|}{\cellcolor[HTML]{B6D7A8}} &
  \multicolumn{1}{l|}{\cellcolor[HTML]{B6D7A8}} &
  \multicolumn{1}{l|}{\cellcolor[HTML]{FFE599}} &
  \multicolumn{1}{l|}{\cellcolor[HTML]{FFE599}} &
   \\ \midrule
\multicolumn{1}{|l|}{\textit{Other}} &
  \multicolumn{1}{l|}{\cellcolor[HTML]{C9DAF8}} &
  \multicolumn{1}{l|}{\cellcolor[HTML]{C9DAF8}} &
  \multicolumn{1}{l|}{\cellcolor[HTML]{C9DAF8}} &
  \multicolumn{1}{c|}{\cellcolor[HTML]{34A853}{0.88}} &
  \multicolumn{1}{l|}{\cellcolor[HTML]{B6D7A8}} &
  \multicolumn{1}{l|}{\cellcolor[HTML]{B6D7A8}} &
  \multicolumn{1}{c|}{\cellcolor[HTML]{34A853}{0.88}} &
  \multicolumn{1}{l|}{\cellcolor[HTML]{FFE599}} &
   \\ \midrule
\multicolumn{10}{|c|}{\cellcolor[HTML]{7F7F7F}{\color[HTML]{FFFFFF} \textbf{Best specificity}}} \\ \midrule
\multicolumn{1}{|l|}{Drinking milk} &
  \multicolumn{1}{l|}{\cellcolor[HTML]{C9DAF8}} &
  \multicolumn{1}{l|}{\cellcolor[HTML]{C9DAF8}} &
  \multicolumn{1}{l|}{\cellcolor[HTML]{C9DAF8}} &
  \multicolumn{1}{c|}{\cellcolor[HTML]{34A853}{0.98}} &
  \multicolumn{1}{l|}{\cellcolor[HTML]{B6D7A8}} &
  \multicolumn{1}{l|}{\cellcolor[HTML]{B6D7A8}} &
  \multicolumn{1}{c|}{\cellcolor[HTML]{34A853}{0.98}} &
  \multicolumn{1}{l|}{\cellcolor[HTML]{FFE599}} &
   \\ \midrule
\multicolumn{1}{|l|}{Grooming} &
  \multicolumn{1}{c|}{\cellcolor[HTML]{34A853}{0.99}} &
  \multicolumn{1}{l|}{\cellcolor[HTML]{C9DAF8}} &
  \multicolumn{1}{l|}{\cellcolor[HTML]{C9DAF8}} &
  \multicolumn{1}{c|}{\cellcolor[HTML]{34A853}{0.99}} &
  \multicolumn{1}{l|}{\cellcolor[HTML]{B6D7A8}} &
  \multicolumn{1}{l|}{\cellcolor[HTML]{B6D7A8}} &
  \multicolumn{1}{c|}{\cellcolor[HTML]{34A853}{0.99}} &
  \multicolumn{1}{l|}{\cellcolor[HTML]{FFE599}} &
   \\ \midrule
\multicolumn{1}{|l|}{Lying} &
  \multicolumn{1}{l|}{\cellcolor[HTML]{C9DAF8}} &
  \multicolumn{1}{c|}{\cellcolor[HTML]{34A853}{0.97}} &
  \multicolumn{1}{l|}{\cellcolor[HTML]{C9DAF8}} &
  \multicolumn{1}{l|}{\cellcolor[HTML]{B6D7A8}} &
  \multicolumn{1}{l|}{\cellcolor[HTML]{B6D7A8}} &
  \multicolumn{1}{l|}{\cellcolor[HTML]{B6D7A8}} &
  \multicolumn{1}{l|}{\cellcolor[HTML]{FFE599}} &
  \multicolumn{1}{c|}{\cellcolor[HTML]{34A853}{0.97}} &
   \\ \midrule
\multicolumn{1}{|l|}{Running} &
  \multicolumn{1}{c|}{\cellcolor[HTML]{34A853}{1.00}} &
  \multicolumn{1}{c|}{\cellcolor[HTML]{34A853}{1.00}} &
  \multicolumn{1}{l|}{\cellcolor[HTML]{C9DAF8}} &
  \multicolumn{1}{c|}{\cellcolor[HTML]{34A853}{1.00}} &
  \multicolumn{1}{c|}{\cellcolor[HTML]{34A853}{1.00}} &
  \multicolumn{1}{l|}{\cellcolor[HTML]{B6D7A8}} &
  \multicolumn{1}{c|}{\cellcolor[HTML]{34A853}{1.00}} &
  \multicolumn{1}{c|}{\cellcolor[HTML]{34A853}{1.00}} &
  \multicolumn{1}{c|}{\cellcolor[HTML]{34A853}{1.00}} \\ \midrule
\multicolumn{1}{|l|}{Walking} &
  \multicolumn{1}{c|}{\cellcolor[HTML]{34A853}{1.00}} &
  \multicolumn{1}{l|}{\cellcolor[HTML]{C9DAF8}} &
  \multicolumn{1}{l|}{\cellcolor[HTML]{C9DAF8}} &
  \multicolumn{1}{c|}{\cellcolor[HTML]{34A853}{1.00}} &
  \multicolumn{1}{l|}{\cellcolor[HTML]{B6D7A8}} &
  \multicolumn{1}{l|}{\cellcolor[HTML]{B6D7A8}} &
  \multicolumn{1}{c|}{\cellcolor[HTML]{34A853}{1.00}} &
  \multicolumn{1}{l|}{\cellcolor[HTML]{FFE599}} &
   \\ \midrule
\multicolumn{1}{|l|}{\textit{Other}} &
  \multicolumn{1}{l|}{\cellcolor[HTML]{C9DAF8}} &
  \multicolumn{1}{l|}{\cellcolor[HTML]{C9DAF8}} &
  \multicolumn{1}{c|}{\cellcolor[HTML]{34A853}{0.94}} &
  \multicolumn{1}{l|}{\cellcolor[HTML]{B6D7A8}} &
  \multicolumn{1}{l|}{\cellcolor[HTML]{B6D7A8}} &
  \multicolumn{1}{l|}{\cellcolor[HTML]{B6D7A8}} &
  \multicolumn{1}{l|}{\cellcolor[HTML]{FFE599}} &
  \multicolumn{1}{l|}{\cellcolor[HTML]{FFE599}} &
   \\ \midrule
\multicolumn{10}{|c|}{\cellcolor[HTML]{7F7F7F}{\color[HTML]{FFFFFF} \textbf{Best precision}}} \\ \midrule
\multicolumn{1}{|l|}{Drinking milk} &
  \multicolumn{1}{l|}{\cellcolor[HTML]{C9DAF8}} &
  \multicolumn{1}{l|}{\cellcolor[HTML]{C9DAF8}} &
  \multicolumn{1}{l|}{\cellcolor[HTML]{C9DAF8}} &
  \multicolumn{1}{l|}{\cellcolor[HTML]{B6D7A8}} &
  \multicolumn{1}{l|}{\cellcolor[HTML]{B6D7A8}} &
  \multicolumn{1}{l|}{\cellcolor[HTML]{B6D7A8}} &
  \multicolumn{1}{c|}{\cellcolor[HTML]{34A853}{0.70}} &
  \multicolumn{1}{l|}{\cellcolor[HTML]{FFE599}} &
   \\ \midrule
\multicolumn{1}{|l|}{Grooming} &
  \multicolumn{1}{l|}{\cellcolor[HTML]{C9DAF8}} &
  \multicolumn{1}{l|}{\cellcolor[HTML]{C9DAF8}} &
  \multicolumn{1}{l|}{\cellcolor[HTML]{C9DAF8}} &
  \multicolumn{1}{c|}{\cellcolor[HTML]{34A853}{0.68}} &
  \multicolumn{1}{l|}{\cellcolor[HTML]{B6D7A8}} &
  \multicolumn{1}{l|}{\cellcolor[HTML]{B6D7A8}} &
  \multicolumn{1}{c|}{\cellcolor[HTML]{34A853}{0.68}} &
  \multicolumn{1}{l|}{\cellcolor[HTML]{FFE599}} &
   \\ \midrule
\multicolumn{1}{|l|}{Lying} &
  \multicolumn{1}{l|}{\cellcolor[HTML]{C9DAF8}} &
  \multicolumn{1}{l|}{\cellcolor[HTML]{C9DAF8}} &
  \multicolumn{1}{l|}{\cellcolor[HTML]{C9DAF8}} &
  \multicolumn{1}{l|}{\cellcolor[HTML]{B6D7A8}} &
  \multicolumn{1}{l|}{\cellcolor[HTML]{B6D7A8}} &
  \multicolumn{1}{l|}{\cellcolor[HTML]{B6D7A8}} &
  \multicolumn{1}{l|}{\cellcolor[HTML]{FFE599}} &
  \multicolumn{1}{c|}{\cellcolor[HTML]{34A853}{0.95}} &
   \\ \midrule
\multicolumn{1}{|l|}{Running} &
  \multicolumn{1}{c|}{\cellcolor[HTML]{34A853}{0.95}} &
  \multicolumn{1}{c|}{\cellcolor[HTML]{34A853}{0.95}} &
  \multicolumn{1}{l|}{\cellcolor[HTML]{C9DAF8}} &
  \multicolumn{1}{c|}{\cellcolor[HTML]{34A853}{0.95}} &
  \multicolumn{1}{l|}{\cellcolor[HTML]{B6D7A8}} &
  \multicolumn{1}{l|}{\cellcolor[HTML]{B6D7A8}} &
  \multicolumn{1}{l|}{\cellcolor[HTML]{FFE599}} &
  \multicolumn{1}{l|}{\cellcolor[HTML]{FFE599}} &
   \\ \midrule
\multicolumn{1}{|l|}{Walking} &
  \multicolumn{1}{l|}{\cellcolor[HTML]{C9DAF8}} &
  \multicolumn{1}{l|}{\cellcolor[HTML]{C9DAF8}} &
  \multicolumn{1}{l|}{\cellcolor[HTML]{C9DAF8}} &
  \multicolumn{1}{c|}{\cellcolor[HTML]{34A853}{0.71}} &
  \multicolumn{1}{l|}{\cellcolor[HTML]{B6D7A8}} &
  \multicolumn{1}{l|}{\cellcolor[HTML]{B6D7A8}} &
  \multicolumn{1}{l|}{\cellcolor[HTML]{FFE599}} &
  \multicolumn{1}{l|}{\cellcolor[HTML]{FFE599}} &
   \\ \midrule
\multicolumn{1}{|l|}{\textit{Other}} &
  \multicolumn{1}{l|}{\cellcolor[HTML]{C9DAF8}} &
  \multicolumn{1}{l|}{\cellcolor[HTML]{C9DAF8}} &
  \multicolumn{1}{l|}{\cellcolor[HTML]{C9DAF8}} &
  \multicolumn{1}{l|}{\cellcolor[HTML]{B6D7A8}} &
  \multicolumn{1}{c|}{\cellcolor[HTML]{34A853}{0.75}} &
  \multicolumn{1}{l|}{\cellcolor[HTML]{B6D7A8}} &
  \multicolumn{1}{l|}{\cellcolor[HTML]{FFE599}} &
  \multicolumn{1}{l|}{\cellcolor[HTML]{FFE599}} &
   \\ \bottomrule
\end{tabular}}
\end{table}

\begin{figure}[hbtp]
\centering
\includegraphics[width=0.95\textwidth]{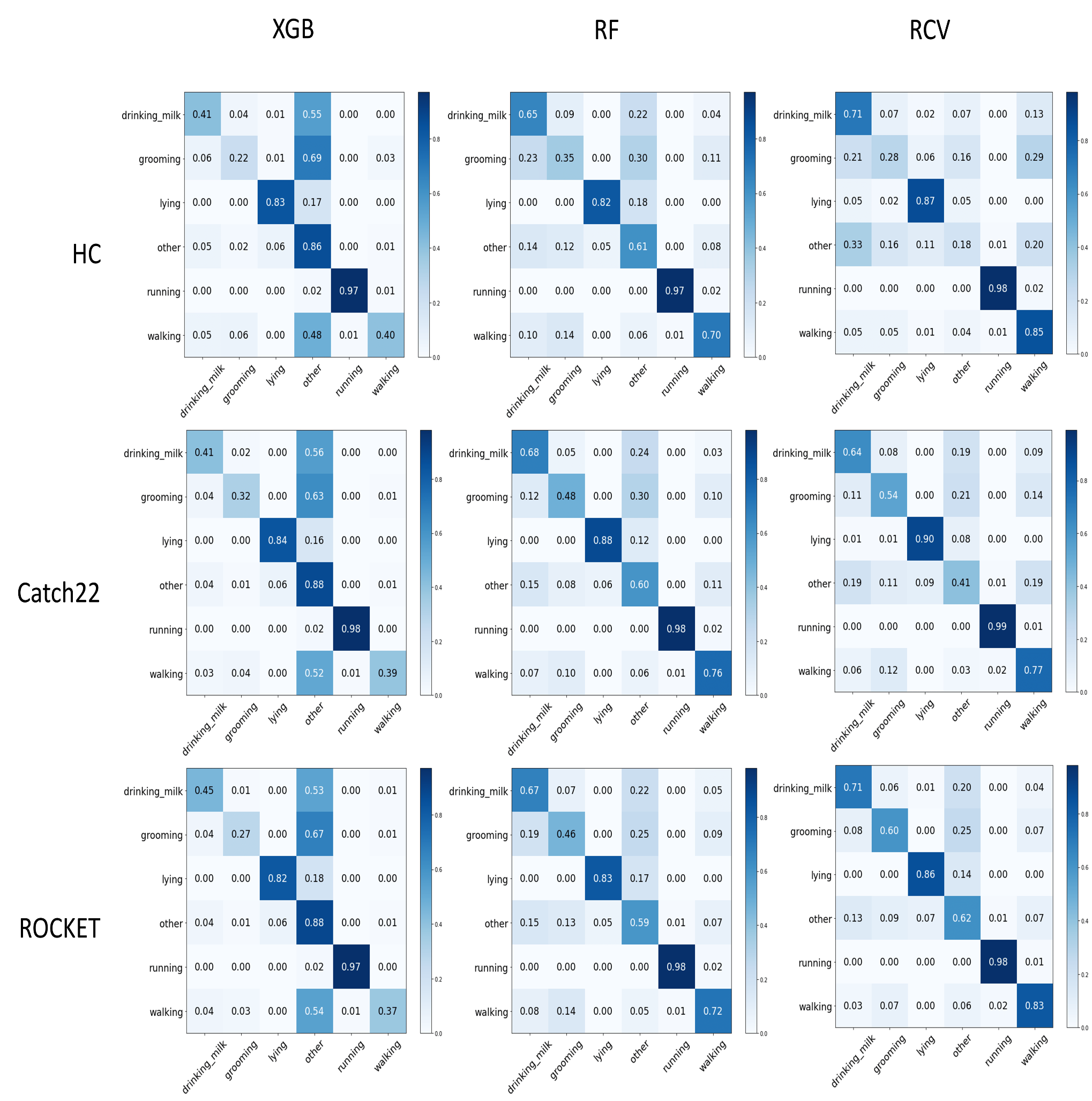}
\captionsetup{labelfont=bf}
\caption{Confusion matrices for each feature-set model combination. Actual behaviours are in row; predicted behaviours are in column. Confusion between behaviours are expressed in percentage.}
\label{fig:cms_results}
\end{figure}

\subsubsection{Calculation time}
\label{sec:calc-time}

Calculation time for features extraction, model training and testing are displayed in Table \ref{tab:time-info}.  While it took 3.8 minutes and 2.05 minutes to extract Catch22 and ROCKET features, respectively, more than 11.5 minutes were necessary for HC feature extraction. However, training of the RCV model was 203 and 64 times longer with ROCKET than with HC and Catch22, respectively. Training XGB model took about five times longer with ROCKET features than with Catch22 and HC, respectively. Model testing took $<1$ second regardless of the feature set-model combination. Based on the total calculation time, including features extraction, model training and testing, the fastest combination was ROCKET with RF (154.06 seconds), which yielded to the third-highest BA (0.71). The highest BA (0.77) obtained from ROCKET with RCV was x3.52 times slower than the fastest feature set-model combination but still took less than 10 minutes.

\begin{table}[hbtp]
\captionsetup{labelfont=bf}
\caption{Calculation time (seconds) for each feature set-model combination, including feature extraction, model training and testing.}
\label{tab:time-info}
\resizebox{\textwidth}{!}{
\begin{tabular}{@{}llccc@{}}
\toprule
                         &                    & XGB       & RandomForest  & RidgeClassifierCV \\ \midrule
\multirow{3}{*}{Hand-Crafted} & Feature Extraction & \multicolumn{3}{c}{697.00} \\
                         & Training           & 491.86    & 42.75         & 2.06              \\
                         & Testing            & 0.16      & 0.42          & 0.01              \\ \midrule
\multirow{3}{*}{Catch22} & Feature Extraction & \multicolumn{3}{c}{228.00}  \\
                         & Training           & 390.51    & 38.64         & 6.54              \\
                         & Testing            & 0.17      & 0.58          & 0.01              \\ \midrule
\multirow{3}{*}{ROCKET}  & Feature Extraction & \multicolumn{3}{c}{123.00} \\
                         & Training           & 2232.42   & 30.47         & 420.00            \\
                         & Testing            & 0.94      & 0.59          & 0.79              \\ \midrule 
\end{tabular}}
\end{table}

\section{Discussion}
\label{sec:discussion}

\subsection{Improving classification performance using ROCKET and Catch22}
\label{sec:disc-1}

The main objective of that study was to evaluate the benefit of ROCKET and Catch22 features to classify pre-weaned calf behaviour from accelerometer data compared to a subset of features typically used in the field. Indeed, ROCKET and Catch22 are explicitly designed for time-series classification problems and exhibit impressive performance in related fields, justifying the evaluation of their performance in livestock ruminant behaviour classification. Therefore, we extracted ROCKET, Catch22 and HC features as the baseline, and tested the performance of the feature sets for the classification of 6 behaviours with 3 Machine Learning models, while considering calf inter-variability in performance evaluation. 
As expected, the best performance was achieved by ROCKET (0.70 ± 0.07 [mean ± std]), closely followed by Catch22 (0.69 ± 0.05), both ranked ahead of HC (0.65 ± 0.03). Especially, the performances of the best combinations of models associated with Catch22 and ROCKET are significantly better than the best combination associated with HC. Indeed, Catch22 features achieved the highest BA of 0.73 using RF, which is a $+5.8\%$ improvement over the highest BA got with HC (0.69). Our study thereby confirms that Catch22 features are promising for classifying livestock ruminant behaviour from accelerometer time-series compared to HC features. Similarly, ROCKET features achieved the highest BA of 0.77 using RCV, which is a $+11.59\%$ improvement over the highest BA got with HC (0.69). Considering that ROCKET can capture local and global patterns in the time-series and achieves better performance than cutting-edge classification techniques \citep{dempster2020rocket}, this finding is consistent with the literature and confirms that ROCKET features must be considered for classifying livestock ruminant behaviour from accelerometer data in further studies.

\subsection{Relevance of feature sets to time-series}
\label{sec:discussion-2}
\begin{figure}[hbtp]
\centering
\includegraphics[width=\textwidth]{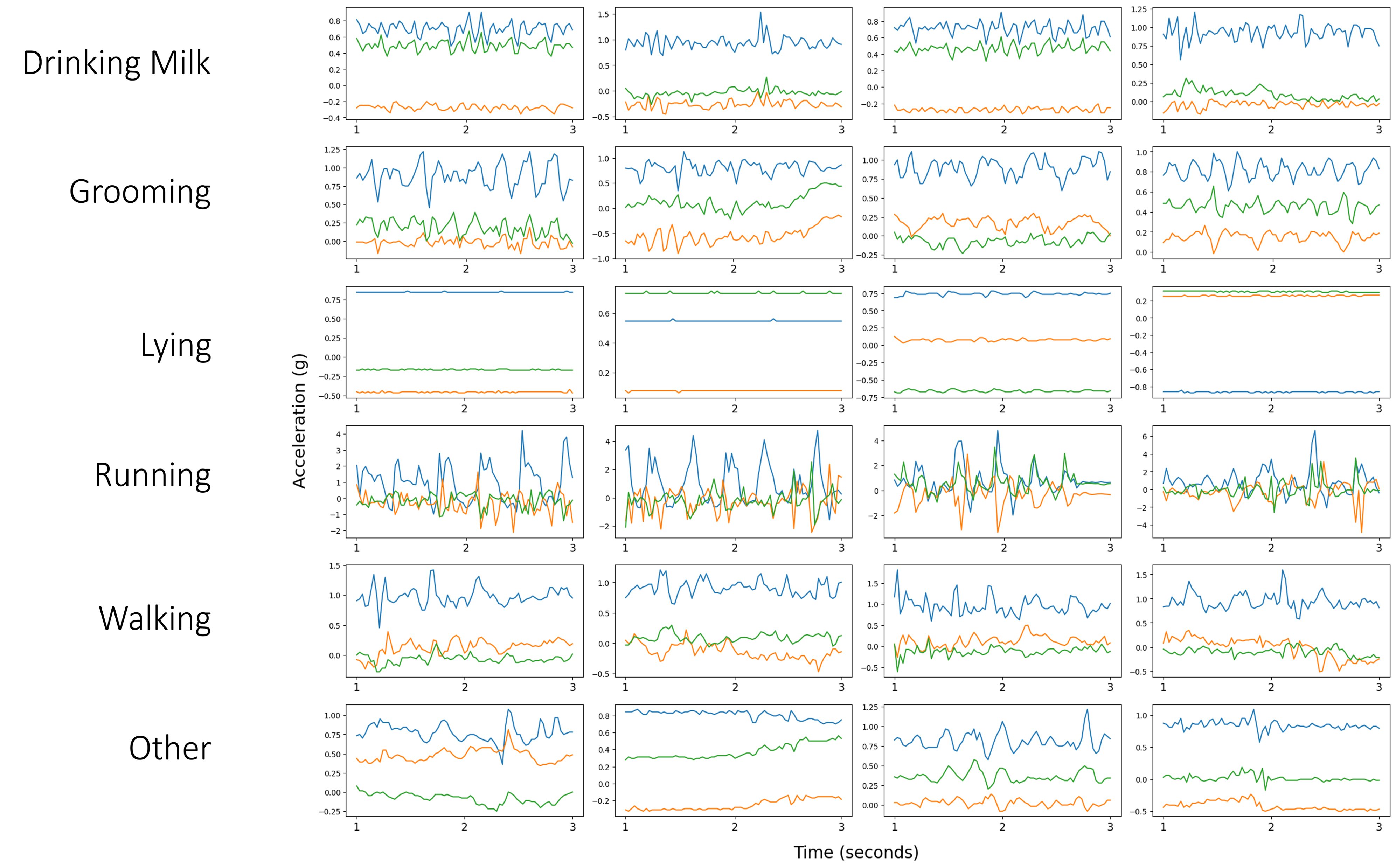}
\captionsetup{labelfont=bf}
\caption{Random 3 seconds time-windows selected from a set of random calves for each behaviour.}
\label{fig:beha_patterns}
\end{figure}

Running and lying performed well with ROCKET, Catch22 and HC features, combining a sensitivity, specificity and precision close to or higher than 0.80 regardless of the Machine Learning model. The singularity of those classes can explain this finding. As illustrated in Figure \ref{fig:beha_patterns}, lying behaviour leads to a flat signal without motion variation. In contrast, running behaviour leads to a high motion intensity signal with substantial variation on the three axes. The time-series signatures in the time-domain are highly specific, explaining why HC, Catch22 and ROCKET features are all able to discriminate lying and running from the \textit{other} behaviours. However, although the specificity is usually higher than 0.80 whatever the behaviour and regardless of the set of features, there is a significant variation in the sensitivity and precision for drinking milk, grooming, walking and \textit{other} depending on the feature set-model combination. HC and Catch22 features are more relevant for classifying walking, especially when combined with RCV and XGB, respectively. The walking time-series do not reveal any apparent shape in the time-domain (Figure \ref{fig:beha_patterns}). This finding is thus consistent with \citet{lubba2019catch22}, who explain that time-series without time-aligned characteristic shapes are better suited for Catch 22 features-based representation. In contrast, ROCKET is more suitable for classifying grooming and drinking milk compared to Catch22 and HC, especially in association with RCV and XGB. As illustrated in Figure \ref{fig:beha_patterns}, grooming and drinking milk time-series have a subtle shape in the time-domain that are almost phase-aligned which is consistent with the temporal structure of these behaviours: Drinking milk is a repetitive sequence starting with milk suckling from the automatic feeder, following by swallowing, while grooming begins with a movement of the head to reach the area of the body to be licked, a sequence of licking movements and a return movement of the head. There is therefore a temporal structure to the movements of the jaw and head which must be reflected in the accelerometer time-series. Finding that ROCKET is more adapted than Catch22 or HC in that context is also coherent with \citet{lubba2019catch22}, who show that shape-based classifier, such as ROCKET and RCV, can accurately capture class differences in the time-series shape. Finally, the 3 features sets are relevant for discriminating the \textit{other} behaviour. This class is the result of merging 19 behaviours, including oral manipulation of the pen, eating concentrates, sniffing, etc. This can be seen in Figure \ref{fig:beha_patterns}, where the time-series of the \textit{other} class in the time-domain look highly different. As a result, some time-series may contain characteristic dynamic properties, while others exhibit subtle shape signatures, which explains why HC, Catch22 and ROCKET actively contribute to the classification of the \textit{other} class. Therefore, our study  confirms that different time-series classification problems require different time-series representations: Classes may be distinguished by multiple types of patterns and dynamical properties \citep{bagnall2017great} and thus complementary features, such as Catch22 and ROCKET, are all necessary to achieve good performance in a multiple tasks classification problem.

\subsection{Impact of the Machine Learning model on the feature sets performance}
\label{sec:discussion-3}
Three Machine Learning models were tested in our study mainly to avoid any bias in evaluating feature sets performance. However, the results highlighted substantial variation within each set of features, suggesting that some Machine Learning models are more adapted to a given set. Based on the BA, the most performant models associated with ROCKET is RCV while the best performant model associated with Catch22 and HC is RF. Those results are consistent with the literature. Indeed, \citet{dempster2020rocket} state that even though ROCKET can be used with any classifier, it is very effective when used in conjunction with linear classifiers that can use a small amount of information from a large number of features. ROCKET and RCV are primarily effective due to their collaboration in managing high-dimensional feature spaces, fast cross-validation for regularization hyperparameters, and regularization mechanisms. This makes them a powerful combination for time-series classification tasks, especially in small datasets \citep{dempster2020rocket, sktime2023rocket, middlehurst2023bake}, as in ruminant behaviour classification, where the volume of data is often limited due to the manual annotation of behaviours (see section \ref{sec:discussion-5}). Finding that RF is the best model associated with Catch22 and HC features is also consistent with the literature \citep{balli2019human, hu2020inclusion, wang2020evaluation, dickinson2021limitations, versluijs2023classification}. RF is especially adequate for handling Catch22 and HC sets due to its ability to handle high-dimensional data and complex feature interactions. Its internal structure, consisting of multiple decision trees, enables robust and accurate modelling of diverse statistical features. Furthermore, even though ROCKET and RCV and Catch22 and RF are the best combinations for overall performance, other combinations look very promising to boost the performance metrics on some behaviours. Especially, Catch22 associated with XGB improves the precision of the behaviour drinking milk (0.69), grooming (0.68) and walking (0.71), which suffer from a large number of false positives with other combinations (see \ref{sec:appendix}). These results show that it is worth testing different models and choosing the one best suited to the features and the configuration of the study, as the best model is undoubtedly highly dependent on the behaviours to be classified, the volume of data, the class balance, etc.

\subsection{Limitations}
\label{sec:discussion-4}
The best overall combination (ROCKET with RCV) reached a BA of 0.77, but improvements are still required to be used for the targeted applications. Also, the best overall combination does not produce the best performance per behaviour when all three metrics (sensitivity, specificity, precision) are considered. As explained above, playing with different feature set-model combinations is necessary to obtain the optimum for each behaviour. In addition, certain combinations can boost the sensitivity of the prediction, but as the expense of the precision. For example, ROCKET with RCV substantially boosted the sensitivity of grooming (0.60) but led to a high level of false positive (precision: 0.40). Similarly, ROCKET with XGB led to the highest precision for drinking milk (0.70) but also led to one of the lowest sensitivities (0.45). In addition, some behaviours lack sensitivity and/or precision, whatever the feature set-model combination (Table \ref{tab:best-per-behav-matrics}). In particular, drinking milk and grooming reached a maximum sensitivity of 0.60 and 0.71, and a precision of 0.68 and 0.70, respectively. Maximum precision of walking was also 0.71. These moderate performances are explained by the confusions between these behaviours (see Figure \ref{fig:cms_results}), probably due to the lack of characteristic dynamic properties and shapes in the accelerometer time-series collected from neck-collars.

These mitigated performances should be seen in light of the highly challenging scenario used in this study. First, six behaviours were kept in the classification to get a sufficient grain to assess calf welfare subsequently. However, the more classes are maintained in the classification, the poorer the performance \citep{riaboff2022predicting}. For example, \citet{martiskainen2009cow} reached an average F1 score of 77.7\% for 8 behaviours-classification using SVM while \citet{benaissa2019use} reached an accuracy of 92\% for 3 behaviours-classification using the same model and accelerometer attachment system. \citet{hosseininoorbin2021deep} observed a 6\% improvement in the average F1 score when reducing the number of classes from 9 to 3. Specifically, the authors showed that performance decreases significantly when more than 5 behaviour classes are considered. This is also consistent with our study, where we observed a 7\% improvement in the BA when decreasing the number of classes from 6 to 4 behaviours using ROCKET and RCV \textit{(data not shown)}.
Furthermore, \citet{hosseininoorbin2021deep} show that walking (F1-score: 32.19), drinking (F1-score: 8.85), grooming (F1-score: 15.60) and \textit{other} (F1-score: 2.03) are extremely difficult to predict despite the robust architecture based on time-frequency data representation combined with Deep Neural Network. Those results are consistent with our study as walking, drinking milk and grooming are particularly difficult to classify, whatever the feature set-model combination. This can be explained by (i) a lack of data available to train and test the model, as these behaviours are rarely observed \citep{arablouei2023animal} and (ii) a lack of characteristic dynamic properties or shapes that do not allow accurate classification. Secondly, the performance in the study were evaluated according to a realistic but challenging scenario where the calves used for model testing have never been used for model training. Indeed, the decrement in model performance linked to high inter-animal variability has been widely reported in the literature \citep{riaboff2022predicting}. This is due to the differences in physical characteristics (e.g., muscles, tendons, joints) that lead to different expressions of the same behaviour (e.g., motion intensity, speed, posture \citep{barwick2020identifying}). To support this, evaluation of ROCKET with RCV with a test set consisting of windows from a random split, irrespective of the calf from which they came, led to a 5\% improvement in BA \textit{(data not shown)}. This trend is notably prominent for grooming and walking, where sensitivity increase (x1.24 and x1.34, respectively) with a random split (test\_size=0.33). This is consistent with \citet{hosseininoorbin2021deep}, observing a 5\% improvement in the F1-score when using random Stratified Cross Validation compared to Leave-One-Animal Cross Validation, with substantial improvement for the most poorly predicted behaviours.

\subsection{Perspective}
\label{sec:discussion-5}
Regarding the limitations, it is worth noting that this study aimed to evaluate the potential of Catch22 and ROCKET features for calf behaviour classification from accelerometer data, considering their high performance for time-series classification in related fields. In that regard, no optimization has been implemented to boost performance. Assuming now that Catch22 and ROCKET are more performant than HC features, several techniques could be implemented to develop an accurate and robust classification system. First, dataset creation could be improved using the best feature set-model combination for each behaviour as a pre-trained model to identify the time-series where less prominent behaviours (drinking, grooming, walking) might occur. This could help to speed up the annotation process while improving the balance between classes, targeting directly the time-series and associated videos where the behaviours of interest may occur rather than annotating videos randomly. This should boost the classifier's performance by enhancing data amount and quality, particularly for grooming and walking behaviours where there is few data available. Moreover, pre-processing could be further developed, investigating the benefit of low-pass filtering to remove signal noise and time-window overlap optimization \citep{riaboff2019evaluation}. Regarding the modelling, feature selection could be applied to select the accelerometer time-series (raw axes, OBDA, pitch, roll, etc.) that are the most informative for that classification task and the features associated with \citep{nanopoulos2001feature}. 

Furthermore, the study must be repeated with more data, different calves in test sets, etc., and if the limitations persist, a multi-vote classifier could be a promising way to improve the results. This multi-vote classifier can be based upon a system like the one suggested by \citet{tahir2012multilabel} where they propose combining existing multi-label methods with ensemble techniques (heterogeneous ensemble). This can also help with sample imbalance and label correlation issues simultaneously. It also closely relates to the stacking idea, where the aggregation process is itself a meta-model (combining multiple machine learning models' predictions with another model enhances accuracy, leveraging the strengths of different base models for improved performance.) \citep{brownlee2021ensemble}.

Another possible system would based on binary classifiers.  \citet{smith2016behavior} implemented a multi-class behaviour modelling based upon the "one-vs-all" framework: A set of binary classifiers was trained to discriminate one of the behaviour classes against all the remaining behaviours merged together. Confidence scores from the binary classifiers were combined to generate the estimated class. This approach led to a 5\% performance improvement over standard multi-class time-series classifiers, creating diversity in the behaviour model. More recently, \citet{arablouei2021situ} also obtained high performance with the "One versus Other" system, where 6 binary classifiers were trained to discriminate the class i versus class j. This approach could be implemented by selecting the best feature-model combinations for each binary classifier. 

Finally, an assumption is that accelerometer data are not informative enough to discriminate behaviours with a fine grain regardless of the set of features. In that scenario, performance could be improved by combining accelerometer data with other sensors \citep{lee2018automatic, cabezas2022analysis, arablouei2023multimodal}. Especially, combining accelerometer data with Computer Vision might help to improve the classification of behaviours which are dependent on the structure of the barn, such as drinking milk, eating concentrates, eating hay, etc. The complementary nature of the information provided by animal tracking  \citep{vayssade2023wizard} should boost substantially the detection of those behaviours while reducing the confusion with the other ones.

\subsection{Practical impact of the study}
\label{sec:discussion-6}

Our study highlighted two sets of features, ROCKET and Catch22, designed explicitly in related communities for time-series classification problems. To the best of the authors' knowledge, Catch22 has never been applied in livestock ruminant classification from accelerometer data, while ROCKET has been implemented just once \citep{brouwers2023towards}. Both ROCKET and Catch22 substantially improved the performance of calf behaviour classification compared to HC features, especially in association with RCV and RF, respectively. Catch22 and ROCKET are also fast to compute and easy to implement using ready-to-use libraries. Therefore, those features are extremely promising for the classification of livestock ruminant behaviour using accelerometer data and should be considered in future studies. In addition, to the authors' knowledge, this is one of the first studies to look at the classification of pre-weaned calf behaviour using accelerometer data, focusing on a wide range of behaviours and testing different methodologies (feature sets and models combinations). Moreover, while some improvement are necessary, performance could be easily optimized, resulting in a system suitable for experimental purposes. As mentioned in section \ref{sec:discussion-5}, a heterogeneous ensemble approach or a binary classifier approach might be effective.

Such a behaviour classification system to monitor drinking milk, running, grooming, walking and lying activities could provide a global view of calf behaviour to support research in calf welfare. That could be used to measure the impact of routine practices (dehorning, transport, weaning from the dams, etc.) on animal behaviour to implement new practices that promote calf welfare in dairy farms. The system could also be applied to quantify the deviance in calf behaviour linked to sickness or distress. Such research may lead to the development of a decision tool to track each calf's behaviours individually and alert farmers to any anomalies detected. Further development would be, however, required to move from a system developed for experimental purposes to a practical system that farmers can use in commercial dairy farms \citep{arablouei2021situ}.

\section{Conclusion}
\label{sec:conclusion}
This study aimed to evaluate the performance of ROCKET and Catch22 features for classifying 6 behaviours in pre-weaned calves, i.e., running, drinking milk, walking, lying, grooming and \textit{other}. For that purpose, the performance of ROCKET and Catch22 features were compared to HC features used in the field along with RF, XGB and RCV Machine Learning models. ROCKET and Catch22 substantially improved the BA compared to HC features. Especially the best feature set-model combination was ROCKET and RCV (BA: 0.77), followed by Catch22 and RF (BA: 0.73) and HC and RF (BA: 0.69). However, ROCKET and RCV do not yield the best sensitivity, specificity and precision for each behaviour. However, some of these variances could disappear with more data, which is for future evaluation. On the contrary, if the results suggest the same conclusions even after more data and optimisations, it can be hypothesised that several feature-set model combinations are necessary to get the optimal prediction of each behaviour separately. This supports the finding that different time-series classification problems require different time-series representations, and thus complementary features, such as Catch22 and ROCKET, are all necessary to perform well in classification. In that regard, a heterogeneous ensemble (multi-vote) system based on the most successful feature set-model combinations or a binary classifiers system would be highly valuable to boost the classification of each behaviour. 
Thus, a behaviour classification system produced as mentioned above could be used to evaluate the effect of practices applied routinely in pre-weaned dairy calves, such as dehorning, transport, weaning from the dams, etc., on calf behaviour to propose recommendations on practices promoting calf welfare in dairy farms. 

\clearpage

\appendix

\section{Results in Detail}
\label{sec:appendix}

\begin{table}[hbtp]
    \begin{center}
            \begin{tabular}{
                @{} >{\columncolor[HTML]{34A853}}l l
                >{\columncolor[HTML]{FF6D01}}l l
                >{\columncolor[HTML]{FFD600}}l l @{}}
                & Best Result & & Worst Result & & Highest mean BA \\
            \end{tabular}
    \end{center}
\end{table}

\begin{table}[!h]
\captionsetup{labelfont=bf}
\caption{Sensitivity values for each feature-model combination per behaviour.}
\label{tab:appendix-sensitivity}
\resizebox{\textwidth}{!}{
\begin{tabular}{@{}|llcccccc|@{}}
\toprule
\multicolumn{8}{|l|}{\cellcolor[HTML]{FFD966}\textbf{Sensitivity}} \\ \midrule
\multicolumn{1}{|l|}{} &
  \multicolumn{1}{l|}{} &
  \multicolumn{1}{c|}{\textbf{drinking\_milk}} &
  \multicolumn{1}{c|}{\textbf{grooming}} &
  \multicolumn{1}{c|}{\textbf{lying}} &
  \multicolumn{1}{c|}{\textbf{running}} &
  \multicolumn{1}{c|}{\textbf{walking}} &
  \textbf{\textit{other}} \\ \midrule
\multicolumn{1}{|l|}{} &
  \multicolumn{1}{l|}{\textbf{XGB}} &
  \multicolumn{1}{c|}{\cellcolor[HTML]{FF6D01}0.41} &
  \multicolumn{1}{c|}{\cellcolor[HTML]{FF6D01}0.22} &
  \multicolumn{1}{c|}{0.83} &
  \multicolumn{1}{c|}{\cellcolor[HTML]{FF6D01}0.97} &
  \multicolumn{1}{c|}{0.40} &
  0.86 \\ \cmidrule(l){2-8} 
\multicolumn{1}{|l|}{} &
  \multicolumn{1}{l|}{\textbf{Random Forest}} &
  \multicolumn{1}{c|}{0.65} &
  \multicolumn{1}{c|}{0.35} &
  \multicolumn{1}{c|}{\cellcolor[HTML]{FF6D01}0.82} &
  \multicolumn{1}{c|}{\cellcolor[HTML]{FF6D01}0.97} &
  \multicolumn{1}{c|}{0.70} &
  0.61 \\ \cmidrule(l){2-8} 
\multicolumn{1}{|l|}{} &
  \multicolumn{1}{l|}{\textbf{RidgeClassifierCV}} &
  \multicolumn{1}{c|}{\cellcolor[HTML]{34A853}0.71} &
  \multicolumn{1}{c|}{0.28} &
  \multicolumn{1}{c|}{0.87} &
  \multicolumn{1}{c|}{0.98} &
  \multicolumn{1}{c|}{\cellcolor[HTML]{34A853}0.85} &
  \cellcolor[HTML]{FF6D01}0.18 \\ \cmidrule(l){2-8} 
\multicolumn{1}{|l|}{} &
  \multicolumn{1}{c|}{\cellcolor[HTML]{D8D8D8}\textit{\textbf{Mean}}} &
  \multicolumn{1}{c|}{\cellcolor[HTML]{D8D8D8}0.59} &
  \multicolumn{1}{c|}{\cellcolor[HTML]{D8D8D8}0.28} &
  \multicolumn{1}{c|}{\cellcolor[HTML]{D8D8D8}0.84} &
  \multicolumn{1}{c|}{\cellcolor[HTML]{D8D8D8}0.98} &
  \multicolumn{1}{c|}{\cellcolor[HTML]{FFFF00}0.65} &
  \cellcolor[HTML]{D8D8D8}0.55 \\ \cmidrule(l){2-8} 
\multicolumn{1}{|l|}{\multirow{-5}{*}{\textbf{Hand-Crafted}}} &
  \multicolumn{1}{c|}{\cellcolor[HTML]{D8D8D8}\textit{\textbf{STD}}} &
  \multicolumn{1}{c|}{\cellcolor[HTML]{D8D8D8}0.16} &
  \multicolumn{1}{c|}{\cellcolor[HTML]{D8D8D8}0.07} &
  \multicolumn{1}{c|}{\cellcolor[HTML]{D8D8D8}0.03} &
  \multicolumn{1}{c|}{\cellcolor[HTML]{D8D8D8}0.01} &
  \multicolumn{1}{c|}{\cellcolor[HTML]{D8D8D8}0.23} &
  \cellcolor[HTML]{D8D8D8}0.34 \\ \midrule
\multicolumn{8}{|l|}{} \\ \midrule
\multicolumn{1}{|l|}{} &
  \multicolumn{1}{l|}{\textbf{XGB}} &
  \multicolumn{1}{c|}{0.42} &
  \multicolumn{1}{c|}{0.32} &
  \multicolumn{1}{c|}{0.84} &
  \multicolumn{1}{c|}{0.98} &
  \multicolumn{1}{c|}{0.40} &
  \cellcolor[HTML]{34A853}0.88 \\ \cmidrule(l){2-8} 
\multicolumn{1}{|l|}{} &
  \multicolumn{1}{l|}{\textbf{Random Forest}} &
  \multicolumn{1}{c|}{0.68} &
  \multicolumn{1}{c|}{0.49} &
  \multicolumn{1}{c|}{0.88} &
  \multicolumn{1}{c|}{0.98} &
  \multicolumn{1}{c|}{0.76} &
  0.61 \\ \cmidrule(l){2-8} 
\multicolumn{1}{|l|}{} &
  \multicolumn{1}{l|}{\textbf{RidgeClassifierCV}} &
  \multicolumn{1}{c|}{0.64} &
  \multicolumn{1}{c|}{0.54} &
  \multicolumn{1}{c|}{\cellcolor[HTML]{34A853}0.90} &
  \multicolumn{1}{c|}{\cellcolor[HTML]{34A853}0.99} &
  \multicolumn{1}{c|}{0.77} &
  0.41 \\ \cmidrule(l){2-8} 
\multicolumn{1}{|l|}{} &
  \multicolumn{1}{c|}{\cellcolor[HTML]{D8D8D8}\textit{\textbf{Mean}}} &
  \multicolumn{1}{c|}{\cellcolor[HTML]{D8D8D8}0.58} &
  \multicolumn{1}{c|}{\cellcolor[HTML]{D8D8D8}0.45} &
  \multicolumn{1}{c|}{\cellcolor[HTML]{FFFF00}0.87} &
  \multicolumn{1}{c|}{\cellcolor[HTML]{D8D8D8}0.98} &
  \multicolumn{1}{c|}{\cellcolor[HTML]{D8D8D8}0.64} &
  \cellcolor[HTML]{D8D8D8}0.63 \\ \cmidrule(l){2-8} 
\multicolumn{1}{|l|}{\multirow{-5}{*}{\textbf{Catch22}}} &
  \multicolumn{1}{c|}{\cellcolor[HTML]{D8D8D8}\textit{\textbf{STD}}} &
  \multicolumn{1}{c|}{\cellcolor[HTML]{D8D8D8}0.14} &
  \multicolumn{1}{c|}{\cellcolor[HTML]{D8D8D8}0.11} &
  \multicolumn{1}{c|}{\cellcolor[HTML]{D8D8D8}0.03} &
  \multicolumn{1}{c|}{\cellcolor[HTML]{D8D8D8}0.00} &
  \multicolumn{1}{c|}{\cellcolor[HTML]{D8D8D8}0.21} &
  \cellcolor[HTML]{D8D8D8}0.24 \\ \midrule
\multicolumn{8}{|l|}{} \\ \midrule
\multicolumn{1}{|l|}{} &
  \multicolumn{1}{l|}{\textbf{XGB}} &
  \multicolumn{1}{c|}{0.45} &
  \multicolumn{1}{c|}{0.27} &
  \multicolumn{1}{c|}{\cellcolor[HTML]{FF6D01}0.82} &
  \multicolumn{1}{c|}{\cellcolor[HTML]{FF6D01}0.97} &
  \multicolumn{1}{c|}{\cellcolor[HTML]{FF6D01}0.38} &
  \cellcolor[HTML]{34A853}0.88 \\ \cmidrule(l){2-8} 
\multicolumn{1}{|l|}{} &
  \multicolumn{1}{l|}{\textbf{Random Forest}} &
  \multicolumn{1}{c|}{0.67} &
  \multicolumn{1}{c|}{0.46} &
  \multicolumn{1}{c|}{0.83} &
  \multicolumn{1}{c|}{0.98} &
  \multicolumn{1}{c|}{0.72} &
  0.59 \\ \cmidrule(l){2-8} 
\multicolumn{1}{|l|}{} &
  \multicolumn{1}{l|}{\textbf{RidgeClassifierCV}} &
  \multicolumn{1}{c|}{\cellcolor[HTML]{34A853}0.71} &
  \multicolumn{1}{c|}{\cellcolor[HTML]{34A853}0.60} &
  \multicolumn{1}{c|}{0.86} &
  \multicolumn{1}{c|}{0.98} &
  \multicolumn{1}{c|}{0.83} &
  0.62 \\ \cmidrule(l){2-8} 
\multicolumn{1}{|l|}{} &
  \multicolumn{1}{c|}{\cellcolor[HTML]{D8D8D8}\textit{\textbf{Mean}}} &
  \multicolumn{1}{c|}{\cellcolor[HTML]{FFFF00}0.61} &
  \multicolumn{1}{c|}{\cellcolor[HTML]{FFFF00}0.45} &
  \multicolumn{1}{c|}{\cellcolor[HTML]{D8D8D8}0.84} &
  \multicolumn{1}{c|}{\cellcolor[HTML]{D8D8D8}0.98} &
  \multicolumn{1}{c|}{\cellcolor[HTML]{D8D8D8}0.64} &
  \cellcolor[HTML]{FFFF00}0.70 \\ \cmidrule(l){2-8} 
\multicolumn{1}{|l|}{\multirow{-5}{*}{\textbf{Rocket}}} &
  \multicolumn{1}{c|}{\cellcolor[HTML]{D8D8D8}\textit{\textbf{STD}}} &
  \multicolumn{1}{c|}{\cellcolor[HTML]{D8D8D8}0.14} &
  \multicolumn{1}{c|}{\cellcolor[HTML]{D8D8D8}0.17} &
  \multicolumn{1}{c|}{\cellcolor[HTML]{D8D8D8}0.02} &
  \multicolumn{1}{c|}{\cellcolor[HTML]{D8D8D8}0.00} &
  \multicolumn{1}{c|}{\cellcolor[HTML]{D8D8D8}0.24} &
  \cellcolor[HTML]{D8D8D8}0.16 \\ \bottomrule
\end{tabular}}
\end{table}

\clearpage

\begin{table}[!h]
\captionsetup{labelfont=bf}
\caption{Specificity values for each feature-model combination per behaviour.}
\label{tab:appendix-specificity}
\resizebox{\textwidth}{!}{
\begin{tabular}{@{}|llcccccc|@{}}
\toprule
\multicolumn{8}{|l|}{\cellcolor[HTML]{FFD966}\textbf{Specificity}} \\ \midrule
\multicolumn{1}{|l|}{} &
  \multicolumn{1}{l|}{} &
  \multicolumn{1}{c|}{\textbf{drinking\_milk}} &
  \multicolumn{1}{c|}{\textbf{grooming}} &
  \multicolumn{1}{c|}{\textbf{lying}} &
  \multicolumn{1}{c|}{\textbf{running}} &
  \multicolumn{1}{c|}{\textbf{walking}} &
  \textbf{\textit{other}} \\ \midrule
\multicolumn{1}{|l|}{} &
  \multicolumn{1}{l|}{\textbf{XGB}} &
  \multicolumn{1}{c|}{0.97} &
  \multicolumn{1}{c|}{\cellcolor[HTML]{34A853}0.99} &
  \multicolumn{1}{c|}{0.96} &
  \multicolumn{1}{c|}{\cellcolor[HTML]{34A853}1.00} &
  \multicolumn{1}{c|}{\cellcolor[HTML]{34A853}1.00} &
  \cellcolor[HTML]{FF6D01}0.72 \\ \cmidrule(l){2-8} 
\multicolumn{1}{|l|}{} &
  \multicolumn{1}{l|}{\textbf{Random Forest}} &
  \multicolumn{1}{c|}{0.92} &
  \multicolumn{1}{c|}{0.94} &
  \multicolumn{1}{c|}{\cellcolor[HTML]{34A853}0.97} &
  \multicolumn{1}{c|}{\cellcolor[HTML]{34A853}1.00} &
  \multicolumn{1}{c|}{0.96} &
  0.82 \\ \cmidrule(l){2-8} 
\multicolumn{1}{|l|}{} &
  \multicolumn{1}{l|}{\textbf{RidgeClassifierCV}} &
  \multicolumn{1}{c|}{\cellcolor[HTML]{FF6D01}0.82} &
  \multicolumn{1}{c|}{\cellcolor[HTML]{FF6D01}0.91} &
  \multicolumn{1}{c|}{\cellcolor[HTML]{FF6D01}0.92} &
  \multicolumn{1}{c|}{1.00} &
  \multicolumn{1}{c|}{\cellcolor[HTML]{FF6D01}0.89} &
  \cellcolor[HTML]{34A853}0.94 \\ \cmidrule(l){2-8} 
\multicolumn{1}{|l|}{} &
  \multicolumn{1}{c|}{\cellcolor[HTML]{D8D8D8}\textit{\textbf{mean}}} &
  \multicolumn{1}{c|}{\cellcolor[HTML]{D8D8D8}0.90} &
  \multicolumn{1}{c|}{\cellcolor[HTML]{D8D8D8}0.95} &
  \multicolumn{1}{c|}{\cellcolor[HTML]{D8D8D8}0.95} &
  \multicolumn{1}{c|}{\cellcolor[HTML]{FFFF00}1.00} &
  \multicolumn{1}{c|}{\cellcolor[HTML]{D8D8D8}0.95} &
  \cellcolor[HTML]{FFFF00}0.83 \\ \cmidrule(l){2-8} 
\multicolumn{1}{|l|}{\multirow{-5}{*}{\textbf{Hand-Crafted}}} &
  \multicolumn{1}{c|}{\cellcolor[HTML]{D8D8D8}\textit{\textbf{STD}}} &
  \multicolumn{1}{c|}{\cellcolor[HTML]{D8D8D8}0.08} &
  \multicolumn{1}{c|}{\cellcolor[HTML]{D8D8D8}0.04} &
  \multicolumn{1}{c|}{\cellcolor[HTML]{D8D8D8}0.03} &
  \multicolumn{1}{c|}{\cellcolor[HTML]{D8D8D8}0.00} &
  \multicolumn{1}{c|}{\cellcolor[HTML]{D8D8D8}0.05} &
  \cellcolor[HTML]{D8D8D8}0.11 \\ \midrule
\multicolumn{8}{|l|}{} \\ \midrule
\multicolumn{1}{|l|}{} &
  \multicolumn{1}{l|}{\textbf{XGB}} &
  \multicolumn{1}{c|}{\cellcolor[HTML]{34A853}0.98} &
  \multicolumn{1}{c|}{\cellcolor[HTML]{34A853}0.99} &
  \multicolumn{1}{c|}{0.96} &
  \multicolumn{1}{c|}{\cellcolor[HTML]{34A853}1.00} &
  \multicolumn{1}{c|}{\cellcolor[HTML]{34A853}1.00} &
  \cellcolor[HTML]{FF6D01}0.72 \\ \cmidrule(l){2-8} 
\multicolumn{1}{|l|}{} &
  \multicolumn{1}{l|}{\textbf{Random Forest}} &
  \multicolumn{1}{c|}{0.92} &
  \multicolumn{1}{c|}{0.96} &
  \multicolumn{1}{c|}{0.96} &
  \multicolumn{1}{c|}{\cellcolor[HTML]{34A853}1.00} &
  \multicolumn{1}{c|}{0.95} &
  0.85 \\ \cmidrule(l){2-8} 
\multicolumn{1}{|l|}{} &
  \multicolumn{1}{l|}{\textbf{RidgeClassifierCV}} &
  \multicolumn{1}{c|}{0.90} &
  \multicolumn{1}{c|}{0.94} &
  \multicolumn{1}{c|}{0.94} &
  \multicolumn{1}{c|}{\cellcolor[HTML]{FF6D01}0.99} &
  \multicolumn{1}{c|}{0.90} &
  0.89 \\ \cmidrule(l){2-8} 
\multicolumn{1}{|l|}{} &
  \multicolumn{1}{c|}{\cellcolor[HTML]{D8D8D8}\textit{\textbf{mean}}} &
  \multicolumn{1}{c|}{\cellcolor[HTML]{D8D8D8}0.93} &
  \multicolumn{1}{c|}{\cellcolor[HTML]{D8D8D8}0.96} &
  \multicolumn{1}{c|}{\cellcolor[HTML]{D8D8D8}0.95} &
  \multicolumn{1}{c|}{\cellcolor[HTML]{FFFF00}1.00} &
  \multicolumn{1}{c|}{\cellcolor[HTML]{D8D8D8}0.95} &
  \cellcolor[HTML]{D8D8D8}0.82 \\ \cmidrule(l){2-8} 
\multicolumn{1}{|l|}{\multirow{-5}{*}{\textbf{Catch22}}} &
  \multicolumn{1}{c|}{\cellcolor[HTML]{D8D8D8}\textit{\textbf{STD}}} &
  \multicolumn{1}{c|}{\cellcolor[HTML]{D8D8D8}0.04} &
  \multicolumn{1}{c|}{\cellcolor[HTML]{D8D8D8}0.03} &
  \multicolumn{1}{c|}{\cellcolor[HTML]{D8D8D8}0.01} &
  \multicolumn{1}{c|}{\cellcolor[HTML]{D8D8D8}0.00} &
  \multicolumn{1}{c|}{\cellcolor[HTML]{D8D8D8}0.05} &
  \cellcolor[HTML]{D8D8D8}0.09 \\ \midrule
\multicolumn{8}{|l|}{} \\ \midrule
\multicolumn{1}{|l|}{} &
  \multicolumn{1}{l|}{\textbf{XGB}} &
  \multicolumn{1}{c|}{\cellcolor[HTML]{34A853}0.98} &
  \multicolumn{1}{c|}{\cellcolor[HTML]{34A853}0.99} &
  \multicolumn{1}{c|}{0.96} &
  \multicolumn{1}{c|}{\cellcolor[HTML]{34A853}1.00} &
  \multicolumn{1}{c|}{\cellcolor[HTML]{34A853}1.00} &
  \cellcolor[HTML]{FF6D01}0.72 \\ \cmidrule(l){2-8} 
\multicolumn{1}{|l|}{} &
  \multicolumn{1}{l|}{\textbf{Random Forest}} &
  \multicolumn{1}{c|}{0.92} &
  \multicolumn{1}{c|}{0.94} &
  \multicolumn{1}{c|}{\cellcolor[HTML]{34A853}0.97} &
  \multicolumn{1}{c|}{\cellcolor[HTML]{34A853}1.00} &
  \multicolumn{1}{c|}{0.96} &
  0.83 \\ \cmidrule(l){2-8} 
\multicolumn{1}{|l|}{} &
  \multicolumn{1}{l|}{\textbf{RidgeClassifierCV}} &
  \multicolumn{1}{c|}{0.93} &
  \multicolumn{1}{c|}{0.95} &
  \multicolumn{1}{c|}{0.95} &
  \multicolumn{1}{c|}{\cellcolor[HTML]{34A853}1.00} &
  \multicolumn{1}{c|}{0.96} &
  0.85 \\ \cmidrule(l){2-8} 
\multicolumn{1}{|l|}{} &
  \multicolumn{1}{c|}{\cellcolor[HTML]{D8D8D8}\textit{\textbf{mean}}} &
  \multicolumn{1}{c|}{\cellcolor[HTML]{FFFF00}0.94} &
  \multicolumn{1}{c|}{\cellcolor[HTML]{FFFF00}0.96} &
  \multicolumn{1}{c|}{\cellcolor[HTML]{FFFF00}0.96} &
  \multicolumn{1}{c|}{\cellcolor[HTML]{FFFF00}1.00} &
  \multicolumn{1}{c|}{\cellcolor[HTML]{FFFF00}0.97} &
  \cellcolor[HTML]{D8D8D8}0.80 \\ \cmidrule(l){2-8} 
\multicolumn{1}{|l|}{\multirow{-5}{*}{\textbf{Rocket}}} &
  \multicolumn{1}{c|}{\cellcolor[HTML]{D8D8D8}\textit{\textbf{STD}}} &
  \multicolumn{1}{c|}{\cellcolor[HTML]{D8D8D8}0.03} &
  \multicolumn{1}{c|}{\cellcolor[HTML]{D8D8D8}0.03} &
  \multicolumn{1}{c|}{\cellcolor[HTML]{D8D8D8}0.01} &
  \multicolumn{1}{c|}{\cellcolor[HTML]{D8D8D8}0.00} &
  \multicolumn{1}{c|}{\cellcolor[HTML]{D8D8D8}0.02} &
  \cellcolor[HTML]{D8D8D8}0.07 \\ \bottomrule
\end{tabular}}
\end{table}

\clearpage

\begin{table}[!h]
\captionsetup{labelfont=bf}
\caption{Precision values for each feature-model combination per behaviour.}
\label{tab:appendix-precision}
\resizebox{\textwidth}{!}{
\begin{tabular}{@{}|llcccccc|@{}}
\toprule
\multicolumn{8}{|l|}{\cellcolor[HTML]{FFD966}\textbf{Precision}} \\ \midrule
\multicolumn{1}{|l|}{} &
  \multicolumn{1}{l|}{} &
  \multicolumn{1}{c|}{\textbf{drinking\_milk}} &
  \multicolumn{1}{c|}{\textbf{grooming}} &
  \multicolumn{1}{c|}{\textbf{lying}} &
  \multicolumn{1}{c|}{\textbf{running}} &
  \multicolumn{1}{c|}{\textbf{walking}} &
  \textbf{\textit{other}} \\ \midrule
\multicolumn{1}{|l|}{} &
  \multicolumn{1}{l|}{\textbf{XGB}} &
  \multicolumn{1}{c|}{0.64} &
  \multicolumn{1}{c|}{0.46} &
  \multicolumn{1}{c|}{0.93} &
  \multicolumn{1}{c|}{\cellcolor[HTML]{34A853}0.95} &
  \multicolumn{1}{c|}{0.59} &
  0.67 \\ \cmidrule(l){2-8} 
\multicolumn{1}{|l|}{} &
  \multicolumn{1}{l|}{\textbf{Random Forest}} &
  \multicolumn{1}{c|}{0.49} &
  \multicolumn{1}{c|}{0.24} &
  \multicolumn{1}{c|}{0.94} &
  \multicolumn{1}{c|}{\cellcolor[HTML]{34A853}0.95} &
  \multicolumn{1}{c|}{0.22} &
  0.69 \\ \cmidrule(l){2-8} 
\multicolumn{1}{|l|}{} &
  \multicolumn{1}{l|}{\textbf{RidgeClassifierCV}} &
  \multicolumn{1}{c|}{\cellcolor[HTML]{FF6D01}0.30} &
  \multicolumn{1}{c|}{\cellcolor[HTML]{FF6D01}0.15} &
  \multicolumn{1}{c|}{\cellcolor[HTML]{FF6D01}0.88} &
  \multicolumn{1}{c|}{0.90} &
  \multicolumn{1}{c|}{\cellcolor[HTML]{FF6D01}0.12} &
  \cellcolor[HTML]{FF6D01}0.66 \\ \cmidrule(l){2-8} 
\multicolumn{1}{|l|}{} &
  \multicolumn{1}{c|}{\cellcolor[HTML]{D8D8D8}\textit{\textbf{mean}}} &
  \multicolumn{1}{c|}{\cellcolor[HTML]{D8D8D8}0.48} &
  \multicolumn{1}{c|}{\cellcolor[HTML]{D8D8D8}0.28} &
  \multicolumn{1}{c|}{\cellcolor[HTML]{D8D8D8}0.92} &
  \multicolumn{1}{c|}{\cellcolor[HTML]{FFFF00}0.93} &
  \multicolumn{1}{c|}{\cellcolor[HTML]{D8D8D8}0.31} &
  \cellcolor[HTML]{D8D8D8}0.67 \\ \cmidrule(l){2-8} 
\multicolumn{1}{|l|}{\multirow{-5}{*}{\textbf{Hand-Crafted}}} &
  \multicolumn{1}{c|}{\cellcolor[HTML]{D8D8D8}\textit{\textbf{STD}}} &
  \multicolumn{1}{c|}{\cellcolor[HTML]{D8D8D8}0.17} &
  \multicolumn{1}{c|}{\cellcolor[HTML]{D8D8D8}0.16} &
  \multicolumn{1}{c|}{\cellcolor[HTML]{D8D8D8}0.03} &
  \multicolumn{1}{c|}{\cellcolor[HTML]{D8D8D8}0.03} &
  \multicolumn{1}{c|}{\cellcolor[HTML]{D8D8D8}0.25} &
  \cellcolor[HTML]{D8D8D8}0.02 \\ \midrule
\multicolumn{8}{|l|}{} \\ \midrule
\multicolumn{1}{|l|}{} &
  \multicolumn{1}{l|}{\textbf{XGB}} &
  \multicolumn{1}{c|}{0.69} &
  \multicolumn{1}{c|}{\cellcolor[HTML]{34A853}0.68} &
  \multicolumn{1}{c|}{0.93} &
  \multicolumn{1}{c|}{\cellcolor[HTML]{34A853}0.95} &
  \multicolumn{1}{c|}{\cellcolor[HTML]{34A853}0.71} &
  0.69 \\ \cmidrule(l){2-8} 
\multicolumn{1}{|l|}{} &
  \multicolumn{1}{l|}{\textbf{Random Forest}} &
  \multicolumn{1}{c|}{0.50} &
  \multicolumn{1}{c|}{0.40} &
  \multicolumn{1}{c|}{0.94} &
  \multicolumn{1}{c|}{0.93} &
  \multicolumn{1}{c|}{0.20} &
  \cellcolor[HTML]{34A853}0.75 \\ \cmidrule(l){2-8} 
\multicolumn{1}{|l|}{} &
  \multicolumn{1}{l|}{\textbf{RidgeClassifierCV}} &
  \multicolumn{1}{c|}{0.43} &
  \multicolumn{1}{c|}{0.33} &
  \multicolumn{1}{c|}{0.91} &
  \multicolumn{1}{c|}{\cellcolor[HTML]{FF6D01}0.83} &
  \multicolumn{1}{c|}{0.13} &
  0.73 \\ \cmidrule(l){2-8} 
\multicolumn{1}{|l|}{} &
  \multicolumn{1}{c|}{\cellcolor[HTML]{D8D8D8}\textit{\textbf{mean}}} &
  \multicolumn{1}{c|}{\cellcolor[HTML]{D8D8D8}0.54} &
  \multicolumn{1}{c|}{\cellcolor[HTML]{FFFF00}0.47} &
  \multicolumn{1}{c|}{\cellcolor[HTML]{D8D8D8}0.92} &
  \multicolumn{1}{c|}{\cellcolor[HTML]{D8D8D8}0.90} &
  \multicolumn{1}{c|}{\cellcolor[HTML]{D8D8D8}0.35} &
  \cellcolor[HTML]{FFFF00}0.72 \\ \cmidrule(l){2-8} 
\multicolumn{1}{|l|}{\multirow{-5}{*}{\textbf{Catch22}}} &
  \multicolumn{1}{c|}{\cellcolor[HTML]{D8D8D8}\textit{\textbf{STD}}} &
  \multicolumn{1}{c|}{\cellcolor[HTML]{D8D8D8}0.14} &
  \multicolumn{1}{c|}{\cellcolor[HTML]{D8D8D8}0.19} &
  \multicolumn{1}{c|}{\cellcolor[HTML]{D8D8D8}0.02} &
  \multicolumn{1}{c|}{\cellcolor[HTML]{D8D8D8}0.06} &
  \multicolumn{1}{c|}{\cellcolor[HTML]{D8D8D8}0.32} &
  \cellcolor[HTML]{D8D8D8}0.03 \\ \midrule
\multicolumn{8}{|l|}{} \\ \midrule
\multicolumn{1}{|l|}{} &
  \multicolumn{1}{l|}{\textbf{XGB}} &
  \multicolumn{1}{c|}{\cellcolor[HTML]{34A853}0.70} &
  \multicolumn{1}{c|}{\cellcolor[HTML]{34A853}0.68} &
  \multicolumn{1}{c|}{0.93} &
  \multicolumn{1}{c|}{0.94} &
  \multicolumn{1}{c|}{0.69} &
  0.67 \\ \cmidrule(l){2-8} 
\multicolumn{1}{|l|}{} &
  \multicolumn{1}{l|}{\textbf{Random Forest}} &
  \multicolumn{1}{c|}{0.48} &
  \multicolumn{1}{c|}{0.28} &
  \multicolumn{1}{c|}{\cellcolor[HTML]{34A853}0.95} &
  \multicolumn{1}{c|}{0.91} &
  \multicolumn{1}{c|}{0.25} &
  0.69 \\ \cmidrule(l){2-8} 
\multicolumn{1}{|l|}{} &
  \multicolumn{1}{l|}{\textbf{RidgeClassifierCV}} &
  \multicolumn{1}{c|}{0.55} &
  \multicolumn{1}{c|}{0.40} &
  \multicolumn{1}{c|}{0.93} &
  \multicolumn{1}{c|}{0.88} &
  \multicolumn{1}{c|}{0.28} &
  0.74 \\ \cmidrule(l){2-8} 
\multicolumn{1}{|l|}{} &
  \multicolumn{1}{c|}{\cellcolor[HTML]{D8D8D8}\textit{\textbf{mean}}} &
  \multicolumn{1}{c|}{\cellcolor[HTML]{FFFF00}0.58} &
  \multicolumn{1}{c|}{\cellcolor[HTML]{D8D8D8}0.45} &
  \multicolumn{1}{c|}{\cellcolor[HTML]{FFFF00}0.94} &
  \multicolumn{1}{c|}{\cellcolor[HTML]{D8D8D8}0.91} &
  \multicolumn{1}{c|}{\cellcolor[HTML]{FFFF00}0.41} &
  \cellcolor[HTML]{D8D8D8}0.70 \\ \cmidrule(l){2-8} 
\multicolumn{1}{|l|}{\multirow{-5}{*}{\textbf{Rocket}}} &
  \multicolumn{1}{c|}{\cellcolor[HTML]{D8D8D8}\textit{\textbf{STD}}} &
  \multicolumn{1}{c|}{\cellcolor[HTML]{D8D8D8}0.11} &
  \multicolumn{1}{c|}{\cellcolor[HTML]{D8D8D8}0.21} &
  \multicolumn{1}{c|}{\cellcolor[HTML]{D8D8D8}0.01} &
  \multicolumn{1}{c|}{\cellcolor[HTML]{D8D8D8}0.03} &
  \multicolumn{1}{c|}{\cellcolor[HTML]{D8D8D8}0.25} &
  \cellcolor[HTML]{D8D8D8}0.03 \\ \bottomrule
\end{tabular}}
\end{table}

 \bibliographystyle{elsarticle-harv} 
 \biboptions{authoryear}
 \bibliography{cas-refs}





\end{document}